\documentclass[10pt,twocolumn,letterpaper]{article}

\usepackage[pagenumbers]{cvpr} 

\usepackage{cuted} 

\usepackage{currfile} 

\usepackage{caption} 
\definecolor{cvprblue}{rgb}{0.21,0.49,0.74}
\usepackage[pagebackref,breaklinks,colorlinks,allcolors=cvprblue]{hyperref}
\usepackage{algpseudocode}
\usepackage{amsmath} 
\usepackage{xcolor}  
\usepackage{tcolorbox}
\usepackage[ruled]{algorithm2e} 
\usepackage{xcolor}
\usepackage{booktabs}    
\usepackage{multirow}    
\usepackage{makecell}    
\usepackage{graphicx}    

\SetAlFnt{\small}
\SetAlCapFnt{\small}
\SetAlCapNameFnt{\small}
\SetAlCapHSkip{0pt}
\def\paperID{*****} 
\def\confName{CVPR}
\def\confYear{2026}

\makeatletter
\newcommand{\removelatexerror}{\let\@latex@error\@gobble}
\makeatother

\newcommand\blfootnote[1]{
    \begingroup
    \renewcommand\thefootnote{}\footnote{#1}
    \addtocounter{footnote}{-1}
    \endgroup
}
\title{Disco-LoRA: Disentangled Composition of Content, Style, and Motion for Multi-concept Video Customization}
\author{Xuancheng Xu\textsuperscript{1} \quad 
Gengyun Jia\textsuperscript{1,2} \quad
Bing-Kun Bao\textsuperscript{2,3,$^\dagger$} \\
\textsuperscript{1}Nanjing University of Posts and Telecommunications \\ 
\textsuperscript{2}Hefei University of Technology \quad  
\textsuperscript{3}Peng Cheng Laboratory \\
{\tt\small 2024010131@njupt.edu.cn} \quad
{\tt\small jgengyun@gmail.com} \quad
{\tt\small bingkunbao@hfut.edu.cn} \\
\\
Project page: \url{https://discolora.github.io/}
}

\begin{document}
\twocolumn[{
    \renewcommand\twocolumn[1][]{#1}%
    \vspace{-2em}
    \maketitle
    \begin{center}
        \centering
        \includegraphics[width=1.0\textwidth]{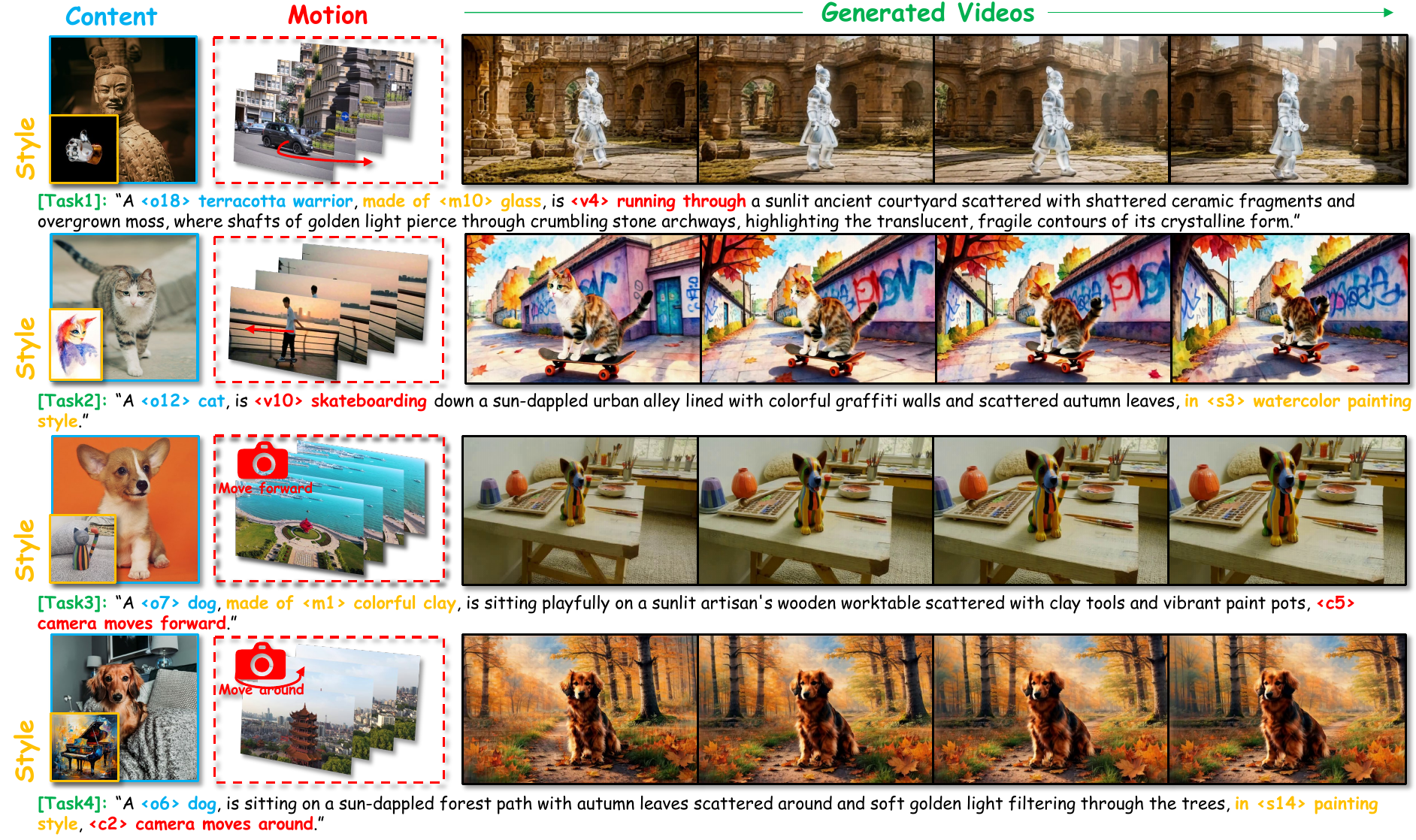}
        \vspace{-2em}
        \captionof{figure}{Disco-LoRA is a customized text-to-video generation framework that enables users to jointly control the object, style, and motion. To evaluate its customization capabilities, we design four distinct tasks, aiming to demonstrate the flexible composition of these disentangled attributes. Specifically, \textbf{Task 1} combines specific object, material texture, and object motion; \textbf{Task 2} integrates the object with an artistic style and object motion; \textbf{Task 3} synthesizes the object, material texture, and camera movement; and \textbf{Task 4} merges the object with artistic style and camera movement.}
        \label{fig_introduction}
        \vspace{-0.5em}
    \end{center}
}]

\blfootnote{$^\dagger$ Corresponding Author}

\maketitle
\begin{abstract}
Video customization based on Text-to-Video (T2V) models aims to learn specific features from reference data to generate controllable videos. 
While significant strides have been made in image stylization and video motion customization, simultaneously controlling multiple concepts, such as content, style, and motion, remains a major challenge.
In this work, we systematically define the task of multi-concept video customization, which requires the joint control of content, style, and motion.
To facilitate research in this area, we construct a comprehensive benchmark and propose Disco-LoRA, a unified framework designed to tackle this problem by disentangling and flexibly recombining different concepts in two stages:
(1) We decompose the objective into two sub-tasks: Content-Style and Content-Motion. Each sub-task is addressed using our Iterative Dual-LoRA Disentanglement Framework, which effectively disentangles distinct concepts within the data.
(2) We identify layer-wise weight trends as crucial for LoRA identity, while weight magnitudes dictate composability. To harmonize these scales, we propose a Z-score-based statistical regularization that aligns weight distributions, preserving layer-wise trends while minimizing interference between different LoRAs.
Extensive experiments show that Disco-LoRA excels in multi-concept video customization, effectively preserving appearance, style, and motion for controllable text-to-video generation.
\end{abstract}  
\section{Introduction}  
Video customization leverages T2V models~\cite{blattmann2023stable, hongcogvideo, chen2024videocrafter2} to synthesize videos using learned reference features. This technique pushes the boundaries of controllable video synthesis, balancing temporal consistency with the preservation of specific content and style.

Recent methods~\cite{wei2024dreamvideo, wang2025dualreal} primarily focus on preserving content appearance from reference images and motion patterns from reference videos, whereas other approaches like~\cite{shah2024ziplora,roy2025duolora} only explore the combination of content and style within static images. However, the customized generation of multi-concept videos that integrates specific content, style, and motion remains a challenging and largely underexplored task.
For instance, designers might need to simulate specific objects interacting dynamically with various styles or materials. 
Likewise, artists may seek to bring specific objects to life within videos that faithfully reflect their own distinctive artistic styles.
Thus, we are the first to systematically define multi-concept video customization as the integration of user-provided content, style, and motion. 
Specifically, we characterize Content by objects, refine Style into material and artistic style based on visual impact, and categorize Motion into object motion and camera movements. 
Accordingly, we formulate four tasks representing all permutations of Content, Style, and Motion, as illustrated in Fig.~\ref{fig_introduction}, and concurrently propose a novel benchmark for multi-concept video customization.

\begin{figure}[t]
    \centering
    \includegraphics[width=0.4\textwidth]{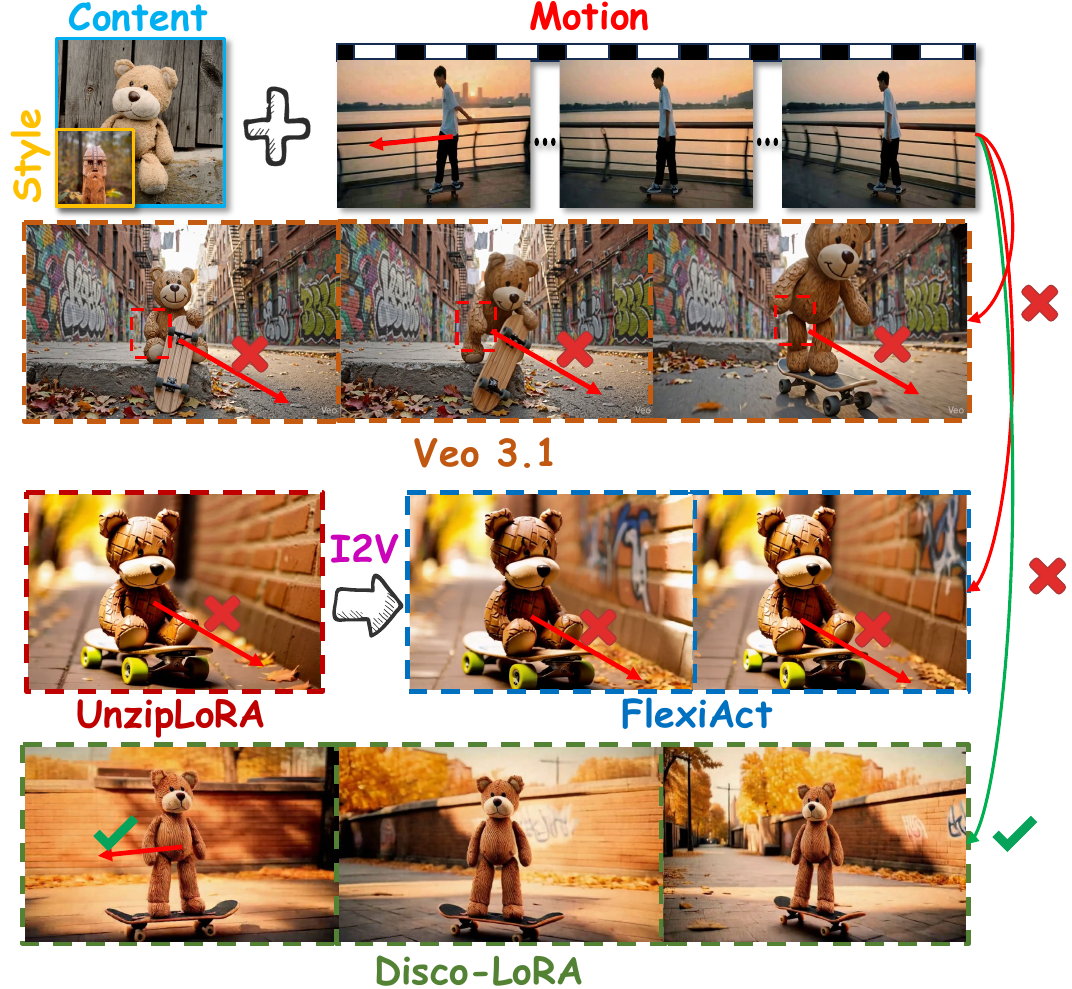}
    \vspace{-1em}
    \caption{Existing proprietary models struggle to maintain material and motion consistency. Furthermore, style-customized initial frames often yield unpredictable layouts, causing subsequent I2V customization to fail. In contrast, Disco-LoRA enables the flexible and precise composition of multiple concepts.}
    \label{fig_introduction_2}
    \vspace{-0.62cm}
\end{figure}

To realize multi-concept video customization, the most intuitive approach is to combine image-based Content-Style customization methods~\cite{shah2024ziplora,liu2025unziplora} with Image-to-Video (I2V) motion customization models~\cite{zhang2025flexiact} to enforce specific motion patterns. 
However, while image-based methods can successfully combine style and content, the position and pose of the generated subjects often remain uncontrollable, as illustrated in Fig.~\ref{fig_introduction_2}, leading to failures in achieving the desired motion customization effects when such images are used as starting frames.
Moreover, recent work ~\cite{huang2025videomage,Veo3} can only achieve multi-subject generation and fails to extend from specific subjects to broader concepts, which limits its overall significance.
To address this limitation, we propose Disco-LoRA, a video customization framework capable of disentangling Content, Style, and Motion, allowing for their arbitrary and free combination.

We identify two core challenges in multi-concept video customization: first, obtaining disentangled representations for each individual concept; and second, combining these representations without mutual interference. 
To address the first challenge, we decompose the complex objective into two independent sub-tasks: Content-Style and Content-Motion disentanglement, which eliminates the need for paired triplets, enabling flexible learning from arbitrary unpaired data. 
Each sub-task is achieved by an Iterative Dual-LoRA Disentanglement Framework that employs iterative learning combined with complementary prompting and time-aware masking, effectively preventing the LoRAs from overfitting to global features, ensuring the separation of content, style, and motion.

Although individual concepts can be disentangled, composing them from disparate sources is hindered by inter-concept interference. Naive combinations typically result in content dominance, where style attributes are overshadowed. 
Investigating this, we identify that layer-wise weight trends are crucial for LoRA identity, whereas weight magnitudes determine composability. 
Our analysis shows that content weights are disproportionately large, disrupting this balance. 
Leveraging this insight, we introduce Z-Score-Based Statistical Regularization. 
By aligning weight distributions through Z-score normalization, we harmonize the magnitudes to ensure composability while preserving the layer-wise trends essential for identity, resulting in a harmonious multi-concept integration.
\begin{figure*}[!h]
    \centering
    \includegraphics[width=0.88\textwidth]{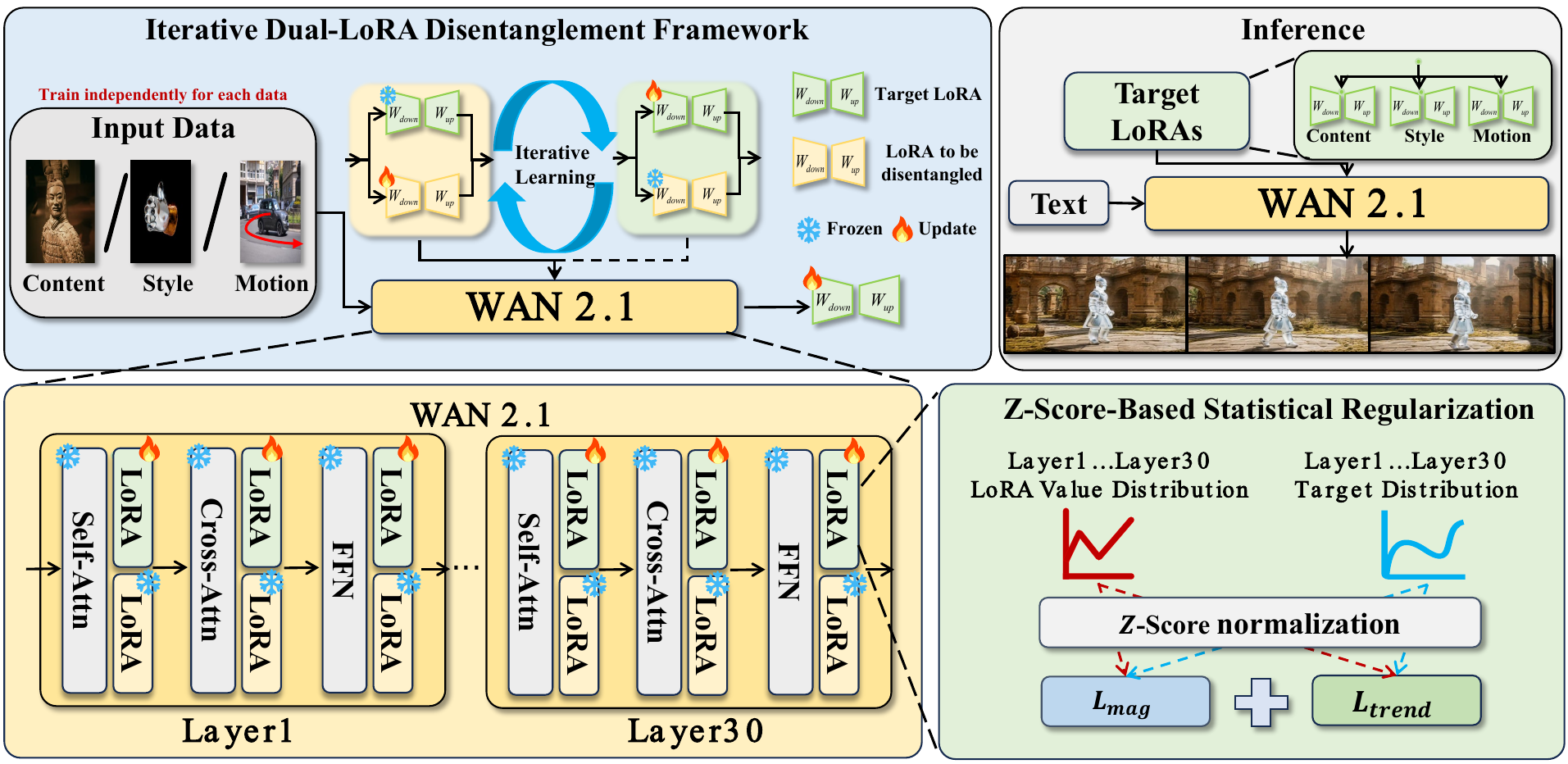}
    \vspace{-0.5pt}
    \caption{\textbf{Overview of Disco-LoRA.} We independently train Content, Style, and Motion using our Iterative Dual-LoRA Disentanglement Framework. We simultaneously train a Target LoRA alongside a LoRA to be disentangled for each data, utilizing the Target LoRA for the final output. Furthermore, we apply Z-Score-Based Statistical Regularization to constrain parameter distributions and prevent concept bleeding. This design realizes free-form multi-concept video customization during inference.}
    \label{fig_pipeline_dit}
    \vspace{-0.2cm}
\end{figure*}
Overall, our contributions can be summarized as:
\begin{itemize}
    \item \textcolor{black}{We define the task of multi-concept video customization by systematically categorizing Content, Style, and Motion, and establish a comprehensive benchmark with four distinct tasks to evaluate this capability.}
\item \textcolor{black}{We introduce an Iterative Dual-LoRA Disentanglement Framework, effectively preventing global overfitting while ensuring robust concept separation.}
\item \textcolor{black}{We propose a Z-score-based regularization method that balances LoRA weights to address magnitude discrepancies, ensuring harmonious concept composition without compromising the specific characteristics of each LoRA.}
\item \textcolor{black}{Our framework enables multi-concept video customization across Content, Style, and Motion, achieving state-of-the-art fidelity in appearance preservation, artistic style transfer, and motion consistency.}
\end{itemize}

\section{Related Works}

\subsection{Text-to-Video Diffusion Transformer}
Diffusion models have achieved remarkable success in generating high-quality videos from text~\cite{blattmann2023stable, singermake, khachatryan2023text2video}. Leveraging the scalability of the Diffusion Transformer (DiT) architecture~\cite{peebles2023scalable}, recent models like~\cite{yang2024cogvideox, wan2025wan} have significantly improved both appearance and motion quality. The proliferation of open-source models has further accelerated this field, enabling realistic video creation.

\subsection{Customized Image Generation}
Customized generation, built upon advances in image generation techniques~\cite{rombach2022high,batifol2025flux,zhang2026meta,zhu2025kv,ma2026group,xu2026processmaker,xu2025chain,xu2024cookgalip,zhang2025ar,zhang2025tar3d}, aims to integrate specific subjects and styles into new contexts while maintaining fidelity~\cite{li2025comprehensive, wang2025stableidentity}. Approaches like~\cite{ruiz2023dreambooth, galimage, 11209272, wu2024infinite} utilized special tokens to anchor subject identity.
However, combining multiple concepts (e.g., subject and style) remains challenging due to feature entanglement. Recent studies~\cite{ jeong2024visual,wang2024instantstyle,cohen2025conditional} address this by analyzing the SDXL~\cite{podell2023sdxl} or jointly training LoRAs with decoupling losses~\cite{shah2024ziplora, frenkel2024implicit, ouyang2025k, roy2025duolora, liu2025unziplora, yang2025qr}. Alternatively, methods such as~\cite{xu2025b4m, po2024orthogonal, zhang2023prospect} learn orthogonal representations for each concept, allowing for interference-free combinatorial generation.

\subsection{Customized Video Generation}
Video customization~\cite{gu2023mix,ma2026fastvmt,ma2025followyourmotion,lireactid,liu2025javisdit} extends image-based customization techniques to the temporal domain and has emerged as an important research direction in controllable generation~\cite{wang2025cinemaster,wang2025characterfactory,wang2026multishotmaster,li2025vfxmaster,ma2025controllable,ma2025followyourclick,ma2024followyouremoji,ma2025followfaster,ma2025followcreation,ma2024followpose,zhu2025memorize}.
While initial works~\cite{jiang2024videobooth, wumotionbooth, wu2025customcrafter, huang2025videomage} focused primarily on static subject fidelity, the greater challenge lies in jointly customizing subjects and motion. SAVE~\cite{song2024save} explores motion personalization for structure-agnostic protagonist editing, enabling a new subject with substantially different body structure to inherit the motion from a single source video.
Approaches that train these modules separately~\cite{wumotionbooth, yang2024direct, wei2024dreamvideo, zhao2024motiondirector, xu2026smrabooth} often suffer from modal interference during inference. Conversely, joint training methods~\cite{wang2025dualreal, chen2025jointtuner} learn appearance and motion simultaneously using paired data. Although effective, this one-to-one paradigm restricts the model's ability to generalize to novel combinations.

\section{Method}
This section introduces Disco-LoRA for multi-concept video customization (Figure~\ref{fig_pipeline_dit}).
Sec.~\ref{sec_dataset} establishes a benchmark dividing the problem into four tasks with specific test prompts.
Sec.~\ref{sec_iterative} details our Iterative Dual-LoRA Disentanglement Framework to ensure robust concept separation.
Finally, Sec.~\ref{sec_zscore} proposes Z-Score Regularization, which preserves layer-wise trends and aligns magnitudes to minimize inter-LoRA interference.
\subsection{Preliminary}\label{sec_Preliminary}
\textbf{Video Diffusion Transformer Model.}  
Text-to-video diffusion transformers utilizing flow matching~\cite{lipman2022flow} have achieved high-quality generation. WAN2.1~\cite{wan2025wan} adopts the strategy, employing WAN block that integrates self-attention for spatiotemporal modeling and cross-attention for text conditioning $P$. The model iteratively denoises Gaussian noise and is optimized via a velocity prediction loss:
\begin{equation}\label{eqn-1}
\mathcal{L}_{diff} = \mathbb{E}_{z_0, z_1, P, t} \left\| u(z_t, P, t; \theta) - v_t \right\|^2,
\end{equation}

\noindent\textbf{Low-Rank Adaptation (LoRA).} LoRA~\cite{hulora} achieves efficient fine-tuning by assuming weight updates $\Delta W$ have a low intrinsic rank. It decomposes the update into low-rank matrices $B \in \mathbb{R}^{m \times r}$ and $A \in \mathbb{R}^{r \times n}$ ($r \ll \min(m, n)$), formulating the forward pass as $W = W_0 + BA$. While $W_0$ remains frozen, only $A$ and $B$ are optimized. When integrating multiple LoRAs, their individual updates are linearly superimposed:
\begin{equation}\label{eqn-1}
W = W_0 + \sum_{i=1}^{N} \Delta W_i,
\end{equation}
where $W_0$ denotes the frozen pre-trained weights, and $\Delta W_i$ represents the weight update contributed by the $i$-th LoRA.

\subsection{Task Definition and Dataset Construction}\label{sec_dataset}
To achieve precise video customization, we establish a rigorous evaluation framework based on the decoupling of three core elements: Content, Style, and Motion.

\noindent\textbf{Task Definition.} We first define Content as the provided visual object element (e.g., a specific plushie teddy bear), which serves as the central entity of the generation. Furthermore, we categorize Style and Motion into sub-types based on their scope of influence. First, we distinguish between \textit{Material Style} and \textit{Artistic Style}. \textit{Material Style} (e.g., ``made of gold'') is object-centric, modifying the subject's texture and physical properties while preserving its geometry. 
In contrast, \textit{Artistic Style} (e.g., ``Van Gogh style'') operates globally, altering the entire video's rendering technique, color palette, and atmosphere. 
Second, we separate \textit{Object Motion} from \textit{Camera Movement}. \textit{Object Motion} refers to the intrinsic dynamics of the object itself (e.g., ``running''), requiring the model to synthesize temporal deformation. Conversely, \textit{Camera Movement} (e.g., ``zoom in'') represents extrinsic observation changes, affecting the viewpoint without necessarily altering the object's state.

Based on these distinctions, we define four tasks to evaluate the model's ability to recombine specific elements: \textcircled{1} Object + Material + Object Motion; \textcircled{2} Object + Artistic Style + Object Motion; \textcircled{3} Object + Material + Camera Movement; and \textcircled{4} Object + Artistic Style + Camera Movement.

\noindent\textbf{Concept Curation.} We curated a diverse dataset comprising 20 content items from DreamBench~\cite{ruiz2023dreambooth}, spanning both dynamic and static categories. Additionally, we selected 10 material and 22 artistic styles from StyleDrop~\cite{sohn2023styledrop}, alongside 10 object motions and 11 camera movements from Davis~\cite{pont20172017} and other online sources. More details are provided in the supplementary material.

\subsection{Iterative Dual-LoRA Disentanglement}\label{sec_iterative}

\noindent\textbf{Task decomposition.} 
Existing multi-concept customization methods face a dilemma: sequential training (e.g., DreamBooth~\cite{ruiz2023dreambooth}) often leads to holistic overfitting, while parallel training (e.g., MotionDirector~\cite{zhao2024motiondirector}) demands strictly paired datasets, imposing high data collection costs. 
To overcome these limitations, we propose to decompose the complex disentanglement objective into two independent sub-tasks based on data modality: \textbf{Content-Style} (from static images) and \textbf{Content-Motion} (from videos). This decoupling obviates the need for paired triplets, allowing efficient learning from arbitrary unpaired data.

\noindent\textbf{Iterative Dual-LoRA Learning.} 
Based on this decomposition, we introduce the Iterative Dual-LoRA Disentanglement Framework. Taking Content-Style disentanglement as an example, we aim to extract distinct content ($\Delta W_c$) and style ($\Delta W_s$) representations from a single image. We model the effective weight $W$ as a superposition of the frozen pre-trained weights $W_0$ and the low-rank updates:
\begin{equation}\label{eqn-3}
W = W_0 + \Delta W_c + \Delta W_s,
\end{equation}
Instead of joint optimization, in each iteration, we update one LoRA while freezing the other (see Algorithm~\ref{alg:iterative_dual_lora} for $\Delta W_c$ ). This forces the active LoRA to capture residual features not encoded by the frozen one, progressively refining the disentanglement.

To prevent information leakage between the two LoRAs, we introduce two regularization strategies:
\textbf{(1) Complementary Prompting Strategy.}
Relying on the same prompts is insufficient to effectively constrain different concepts within a single data sample for separation; thus, we use a complementary prompting strategy.
We first train the Style LoRA with a style-only prompt $P_s=$ ``in <style>''. Subsequently, the Content LoRA is trained using the composite prompt $P_c=$ ``A <content>'' + $P_s$. This formulation establishes $P_s$ as a stylistic basis, compelling the Content LoRA to learn only the remaining semantic content features.
\textbf{(2) Time-aware Masking Strategy.}
To mitigate overfitting to global image in single-image training, we leverage the diffusion process's property where structure is determined early and details later. We implement a time-aware masking strategy (see Alg.~\ref{alg:iterative_dual_lora}, L4 \& L11, utilizing time-aware thresholds $T_l$ and $T_h$) that constrains optimization based on the diffusion timestep. By targeting structural outlines during high-noise steps and fine textures during low-noise steps, we ensure a distinct allocation of features between the two LoRAs.

The proposed framework naturally extends to video data by learning a Content LoRA ($\Delta W_c$) and a Motion LoRA ($\Delta W_m$). While sharing the iterative logic and constraints described above, \(\Delta W_c\) is trained exclusively on the first frame, whereas \(\Delta W_m\) is trained on the entire sequence. Since \(\Delta W_m\) is learned as a residual on top of the frozen static content, it is mathematically forced to model the temporal dynamics, achieving effective motion disentanglement without complex auxiliary losses.

\begin{algorithm}[htb]
\caption{Iterative Dual-LoRA Disentanglement for \textbf{\textcolor{red}{$\Delta W_c$}}}
\label{alg:iterative_dual_lora}
\begin{algorithmic}[1]
\Require Total epochs $\mathcal{N}$
\Require Pre-trained weights $W_0$, training data $\mathcal{D}$
\Require Content LoRA $\Delta W_c$ (Target LoRA) and Style LoRA $\Delta W_s$ (LoRA to be disentangled)
\Require Prompts $P_c$ (e.g., ``A $\langle$content$\rangle$, in $\langle$style$\rangle$'')
\Require Prompts $P_s$ (e.g., ``in $\langle$style$\rangle$'')
\Require Timestep thresholds $T_l, T_h$ and total timesteps $T$, where $0 \leq T_l \leq T_h \leq T$

\State \textbf{for} each training epoch $i \in \mathcal{N}$ \textbf{do}
\Statex \hspace{\algorithmicindent} \textcolor{gray}{\textbf{//Phase 1: Update Style LoRA}}
\State \hspace{\algorithmicindent} \textbf{for} each training step or batch $x \in \mathcal{D}$ \textbf{do}
    \State \hspace{\algorithmicindent}\hspace{1em} \textcolor{red}{\textbf{Update} $\Delta W_s$}; \textcolor{blue}{\textbf{Freeze} $\Delta W_c$}
    \State \hspace{\algorithmicindent}\hspace{1em} Sample timestep $t \sim \mathcal{U}(0, T_l)$
    \State \hspace{\algorithmicindent}\hspace{1em} $W \leftarrow W_0 + \text{StopGrad}(\Delta W_c) + \Delta W_s$
    \State \hspace{\algorithmicindent}\hspace{1em} $\mathcal{L}_s \leftarrow \mathcal{L}_{\text{diff}}(x, P_s, t; W)$
    \State \hspace{\algorithmicindent}\hspace{1em} $\Delta W_s \leftarrow \Delta W_s - \eta \frac{\partial \mathcal{L}_s}{\partial \Delta W_s}$
\State \hspace{\algorithmicindent} \textbf{end for}
    
\Statex \hspace{\algorithmicindent} \textcolor{gray}{\textbf{//Phase 2: Update Content LoRA}}
\State \hspace{\algorithmicindent} \textbf{for} each training step or batch $x \in \mathcal{D}$ \textbf{do}
    \State \hspace{\algorithmicindent}\hspace{1em} \textcolor{blue}{\textbf{Freeze} $\Delta W_s$}; \textcolor{red}{\textbf{Update} $\Delta W_c$}
    \State \hspace{\algorithmicindent}\hspace{1em} Sample timestep $t \sim \mathcal{U}(T_l, T_h)$
    \State \hspace{\algorithmicindent}\hspace{1em} $W \leftarrow W_0 + \Delta W_c + \text{StopGrad}(\Delta W_s)$
    \State \hspace{\algorithmicindent}\hspace{1em} $\mathcal{L}_c \leftarrow \mathcal{L}_{\text{diff}}(x, P_c, t; W)$
    \State \hspace{\algorithmicindent}\hspace{1em} $\Delta W_c \leftarrow \Delta W_c - \eta \frac{\partial \mathcal{L}_c}{\partial \Delta W_c}$
\State \hspace{\algorithmicindent} \textbf{end for}
\State \textbf{end for}

\State \textbf{Return} \textbf{\textcolor{red}{$\Delta W_c$}}
\end{algorithmic}
\end{algorithm}

\subsection{Z-Score-Based Statistical Regularization}\label{sec_zscore}
\begin{figure*}[t]
    \centering
    \includegraphics[width=0.90\textwidth]{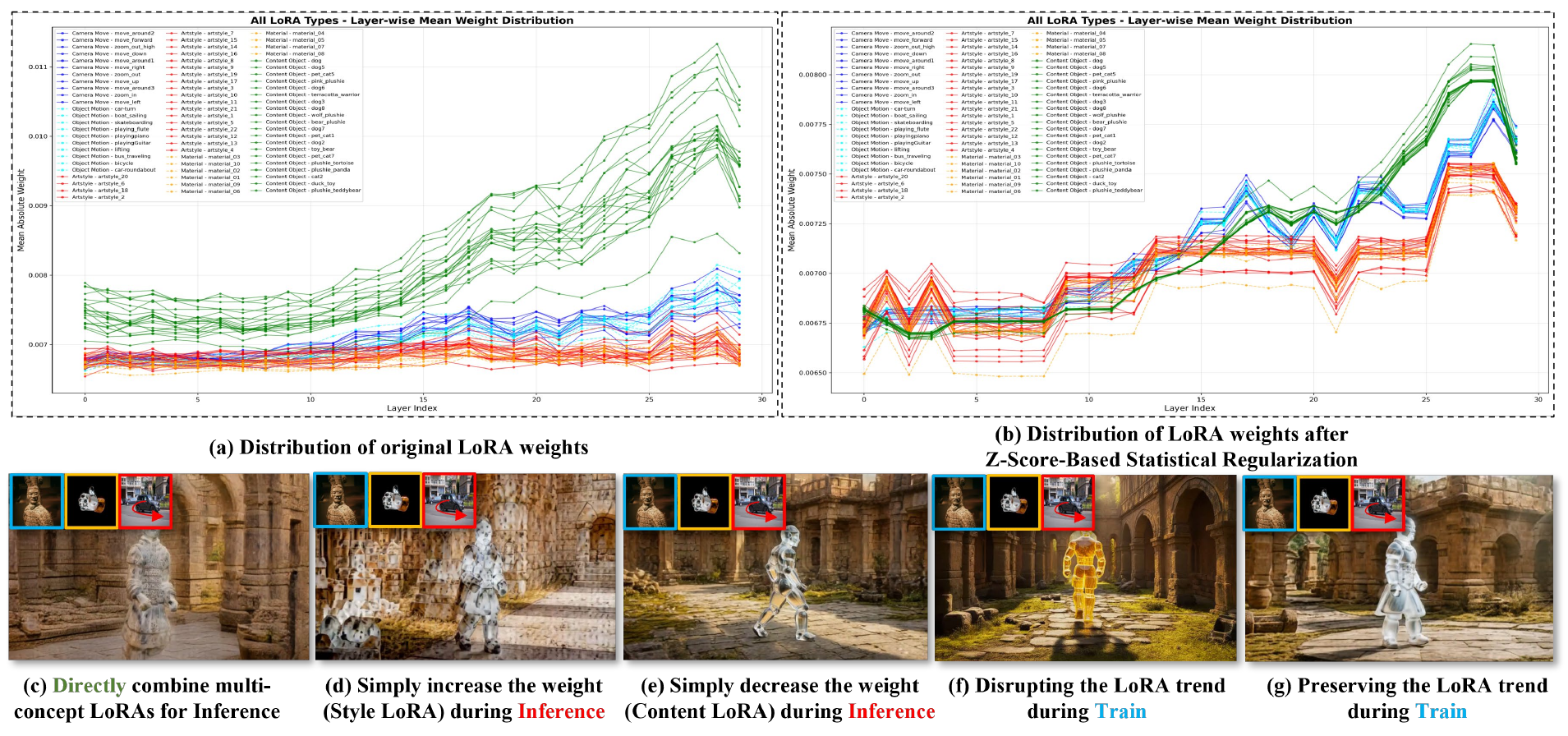}
    \vspace{-5pt}
    \caption{Visual analysis of Z-Score-Based Statistical Regularization. (a) Mean curves of the original Content, Style, and Motion LoRAs. (b) Results after Statistical Regularization: means across concepts are aligned, while original inter-layer trends are preserved. (c) Results after directly combining multi-concept LoRAs for Inference. (d) Naively boosting a low-value LoRA causes severe artifacts. (e) Reducing a high-value Content LoRA degrades subject identity. (f) Altering inter-layer trends harms LoRA performance. (g) Our method achieves optimal composition by aligning distributions without disrupting these trends.}
    \label{fig_curve}
    \vspace{-0.2cm}
\end{figure*}

While our framework effectively disentangles concepts, combining LoRAs from diverse sources remains challenging. Naive linear combination often results in severe interference, particularly content dominance (Fig.\ref{fig_curve}(c)). Visualizing the mean weights across DiT layers reveals the root cause: while LoRAs of the same type share similar layer-wise trends, their magnitudes vary significantly (Fig.\ref{fig_curve}(a)). Notably, Content LoRA weights possess much higher magnitudes than other types. Since LoRAs are combined linearly (Eq.~\ref{eqn-3}), this disparity causes content features to overshadow others.

\begin{figure}[t]
    \centering
    \includegraphics[width=0.45\textwidth]{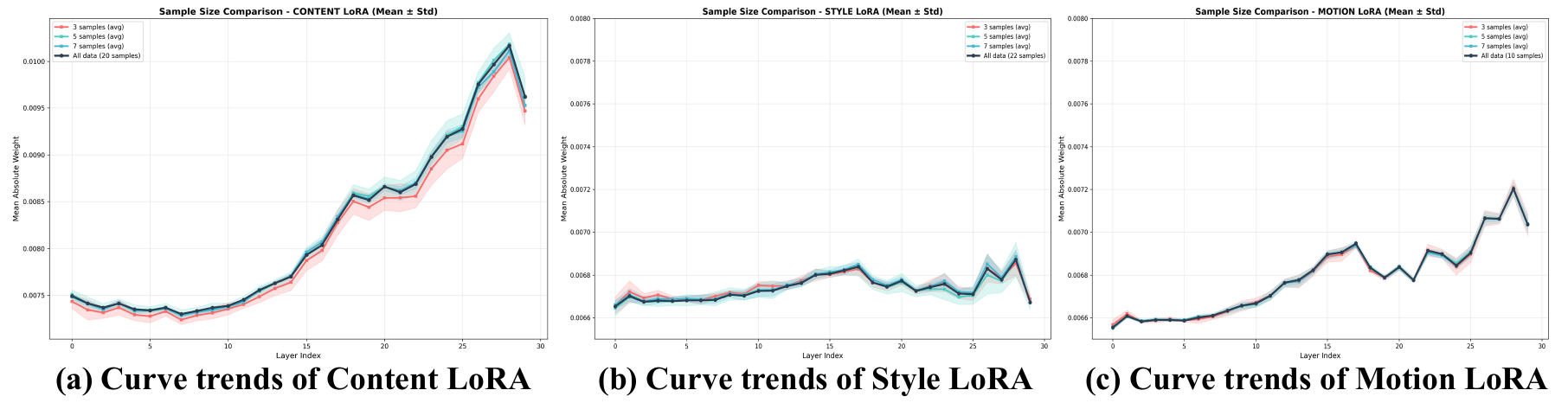}
    \vspace{-5pt}
    \caption{\textbf{Analysis of ground truth selection for three different LoRA concepts.} We demonstrate that the trend of the mean curves remains consistent regardless of the sample size, despite minor differences in value ranges. Consequently, by preserving the trend and adjusting the LoRA value range, we can utilize a curve derived from averaging all cases. This approach is generalizable and robust, remaining applicable even as the dataset size expands in the future.}
    \label{fig_zscore}
\end{figure}
To address this, we seek to align the value ranges of different LoRAs. We find that simple rescaling during inference is ineffective, disrupting the learned weight distributions leads to visual artifacts or identity loss (Fig.\ref{fig_curve}(d, e)). Instead, we propose Z-Score-Based Statistical Regularization. 
This method independently computes curves for the three concepts and constrains their weights to a unified numerical range during training. This ensures magnitude alignment while explicitly preserving the original layer-wise trends essential for accurate feature expression (Fig. \ref{fig_curve}(b)).

We guide the training process by aligning layer-wise weight magnitudes with a target distribution derived from an offline analysis of well-trained LoRA models. Specifically, we shift the overall curves shown in Fig.~\ref{fig_curve}(a) to ensure that the means of the three different data types are approximately consistent. Although these mean curves are fitted using the entire dataset, they exhibit strong generalization capabilities. Therefore, no modifications are required when training on new data.
Let the target vector be $\mathbf{T} = [t_0, \dots, t_{L-1}]$, where each $t_l$ represents the average absolute weight magnitude of layer $l$, $L$ represents the total layer number of WAN 2.1. During training, we compute the current magnitude $c_l$ for layer $l$ by aggregating its constituent LoRA modules $\mathcal{M}_l = {(\mathbf{A}i, \mathbf{B}i)}_{i=1}^{N_l}$:
\begin{equation}
c_l = \frac{1}{2N_l} \sum_{i=1}^{N_l} \left( \overline{|\mathbf{A}_i|} + \overline{|\mathbf{B}_i|} \right),
\end{equation}
where $\overline{|\cdot|}$ denotes the arithmetic mean. We regularize the individual matrices $\mathbf{A}$ and $\mathbf{B}$ rather than their product to prevent numerical instability caused by extremely small values and to maintain consistency with our offline analysis.

To explicitly capture the relative distribution pattern across layers, we employ Z-score normalization. This strategy allows us to isolate the trend of the weight distribution. We normalize both the current magnitudes ${c_l}$ and the target vector ${t_l}$ as follows:
\begin{equation}
z_l^{(c)} = \frac{c_l - \mu_c}{\sigma_c + \epsilon}, \quad z_l^{(t)} = \frac{t_l - \mu_t}{\sigma_t + \epsilon},
\end{equation}
where $\mu$ and $\sigma$ represent the mean and standard deviation of the distributions across all $L$ layers, and $\epsilon=10^{-8}$ ensures stability.

Based on this normalization, we formulate the regularization objective by combining a trend-aware loss with an absolute magnitude constraint. The Trend Loss minimizes the discrepancy between the normalized trends using Mean Squared Error (MSE):
\begin{equation}
\mathcal{L}_{\text{trend}} = \frac{1}{L} \sum_{l=0}^{L-1} \left( z_l^{(c)} - z_l^{(t)} \right)^2,
\end{equation}
Simultaneously, to prevent the weights from drifting in absolute scale, we incorporate a Magnitude Loss using an $L_1$ distance:
\begin{equation}
\mathcal{L}_{\text{mag}} = \frac{1}{L} \sum_{l=0}^{L-1} \left|  c_l - t_l \right|,
\end{equation}
The final regularization term is a weighted sum of these two components, which is integrated into the primary diffusion objective:
\begin{equation}
\mathcal{L} = \mathcal{L}_{\text{diff}} + \lambda_{\text{mag}}\mathcal{L}_{\text{mag}} + \lambda_{\text{trend}} \mathcal{L}_{\text{trend}}.
\end{equation}

\begin{table*}[t]
\centering
\footnotesize
\renewcommand{\arraystretch}{0.9} 
\setlength{\tabcolsep}{1.5pt} 

\caption{Quantitative experimental results for different DiT-based methods under the numerical evaluation metrics.}
\vspace{-1em}

\begin{tabular}{@{} l *{11}{c} @{}}
\toprule
\multirow{2}{*}[-1.7ex]{Method} & \multicolumn{5}{c}{\textbf{Semantic Alignment}} & \multicolumn{3}{c}{\textbf{Motion Quality}} & \multicolumn{3}{c}{\textbf{Perceptual Quality}} \\ 

\cmidrule(r){2-6} \cmidrule(lr){7-9} \cmidrule(l){10-12}

& \makecell[c]{CLIP-T}\textcolor{red}{$\uparrow$} 
& \makecell[c]{CLIP-I(S)}\textcolor{red}{$\uparrow$} 
& \makecell[c]{CLIP-I(C)}\textcolor{red}{$\uparrow$} 
& \makecell[c]{CLIP-I(A)}\textcolor{red}{$\uparrow$} 
& \makecell[c]{CSD}\textcolor{red}{$\uparrow$} 
& \makecell[c]{Motion\\Fidelity}\textcolor{red}{$\uparrow$} 
& \makecell[c]{Subject\\Consistency}\textcolor{red}{$\uparrow$} 
& \makecell[c]{Motion\\Smooth}\textcolor{red}{$\uparrow$} 
& \makecell[c]{Pick\\Score}\textcolor{red}{$\uparrow$} 
& \makecell[c]{Aesthetic\\Quality}\textcolor{red}{$\uparrow$}  
& \makecell[c]{Imaging\\Quality}\textcolor{red}{$\uparrow$} 
\\ \midrule  

DreamBooth(WAN)  & 0.298 & 0.522 & \underline{0.695} & 0.609 & 0.151 & 0.687 & 0.970 & 0.986  & 0.198 & 0.624 & 0.668 \\ 
MotionDirector(WAN) & 0.308 & 0.541 & \textbf{0.698} & 0.620 & 0.179 & 0.701 & \textbf{0.976} & \textbf{0.987} & 0.200  & 0.629  & 0.680 \\ 
UnzipLoRA+FlexiAct & 0.348 & 0.583 & 0.643  & 0.613 & 0.204 & 0.491 & 0.942 & 0.982 & 0.209  & 0.648  & 0.681 \\ 
\textbf{Disco-LoRA (Ours)} & \textbf{0.361} & \textbf{0.592} & 0.673 & \textbf{0.633} & \textbf{0.211} & \textbf{0.771} & \underline{0.971} & \textbf{0.987}  & \textbf{0.212}  & \textbf{0.683}  & \textbf{0.694}\\ 
\midrule
w/o. C.P   & 0.315 & 0.530 & 0.635 & 0.583 & 0.187 & 0.695 & 0.945 & 0.977 & 0.201 & 0.635 & 0.675 \\ 
w/o. T.A.M & 0.345 & \textbf{0.601} & 0.640 & \underline{0.621} & \underline{0.210} & \underline{0.768} & 0.965 & \textbf{0.988} & \underline{0.208} & \underline{0.660} & \underline{0.688} \\ 
w/o. $L_{trend}$ & 0.310 & 0.515 & 0.610 & 0.563 & 0.181 & 0.580 & 0.920 & 0.975 & 0.201 & 0.620 & 0.665 \\ 
w/o. $L_{mag}$ & 0.330 & 0.545 & \textbf{0.684} & 0.615 & 0.196 & 0.725 & \textbf{0.972} & 0.983 & 0.205 & 0.640 & 0.682 \\ 
\textbf{Disco-LoRA (Ours)} & \textbf{0.361} & \underline{0.592} & \underline{0.673} & \textbf{0.633} & \textbf{0.211} & \textbf{0.771} & \underline{0.971} & \underline{0.987}  & \textbf{0.212}  & \textbf{0.683}  & \textbf{0.694}\\ 
\bottomrule
\end{tabular}
\label{table_comparison}
\vspace{-0.3cm}
\end{table*}
\begin{figure*}[t]
    \centering
    \includegraphics[width=0.85\textwidth]{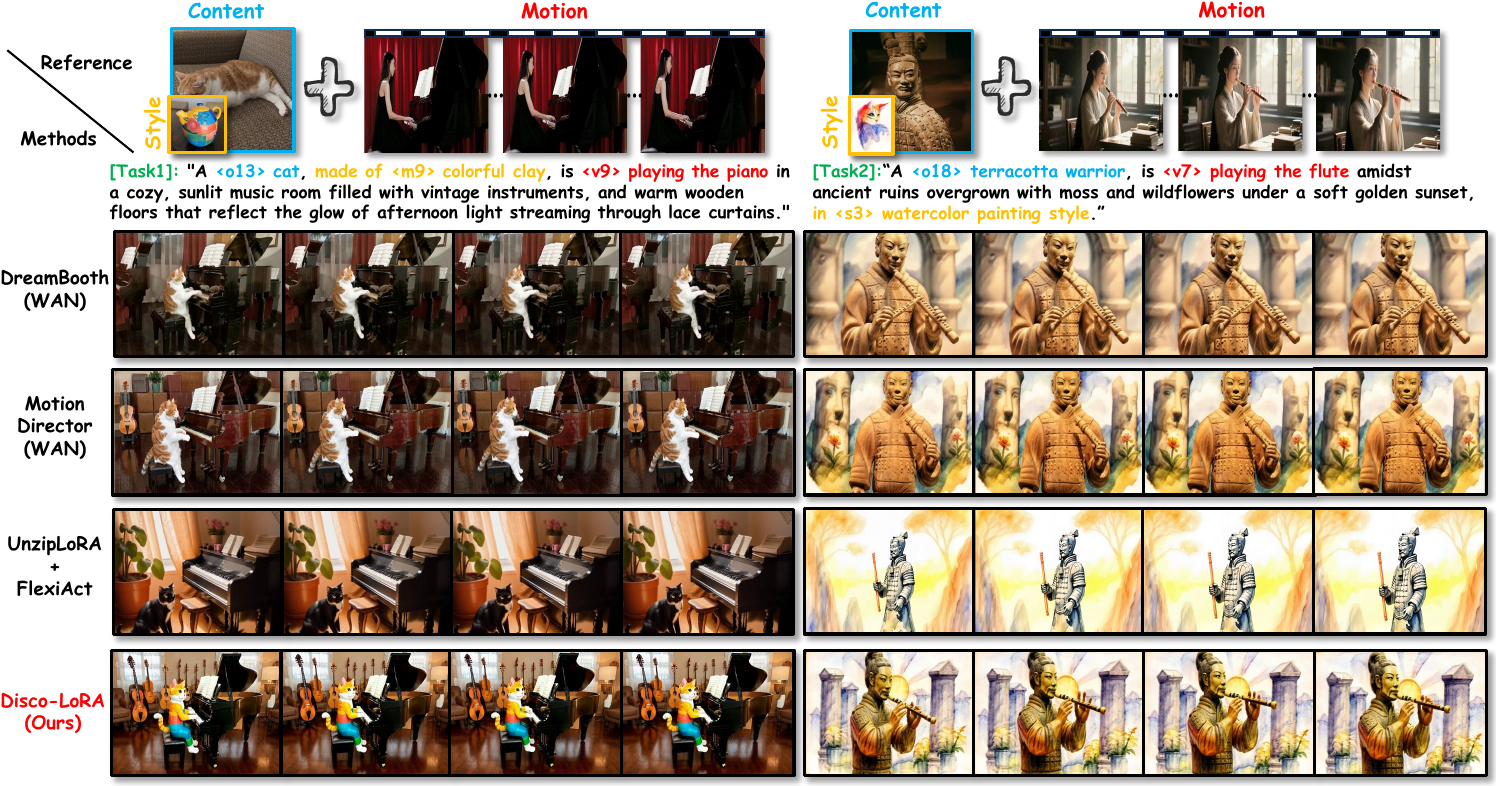}
    \vspace{-10pt}
    \caption{\textbf{Qualitative comparison of Multi-concept Video Customizatio for Task1 and Task2.} Disco-LoRA preserves content identity, style similarity and object motion patterns, while other methods fail to stay faithful to the reference.}
    \label{fig_qua1}
    \vspace{-0.25cm}
\end{figure*}
Furthermore, we analyze the robustness of these curves in Fig.~\ref{fig_zscore}. For each independent concept, we validate the trend stability using smaller subsets of data (e.g., 3, 5, and 7 cases). Notably, despite the reduced dataset size, the trend of the curve derived from the mean remains fundamentally unchanged, exhibiting only minor fluctuations in magnitude. Consequently, we conclude that the curve's trend is independent of dataset quantity. The observed trend possesses significant generalization capabilities; thus, as datasets expand in the future, our method can generalize to new data without requiring re-adjustment of the curve distribution.

\section{Experiment}

\subsection{Experimental Settings}
\noindent\textbf{\emph{Implementation Details.}} For Content and Style learning, we train the LoRA with a learning rate of $1.0 \times 10^{-4}$ and the rank of 32.
For motion learning, the LoRA is trained with the same learning rate and a rank of 64, with videos sampled to 49 frames at a resolution of \(576 \times 320\). 
During inference, we use a 50-step DDIM sampler~\cite{songdenoising} and classifier-free guidance~\cite{ho2022classifier} to generate 49-frame videos at 15 fps and $832 \times 480$ resolution. 
Experiments are conducted on two 96G NVIDIA Pro-6000 GPUs using the DiT~\cite{peebles2023scalable}-based WAN2.1~\cite{wan2025wan} 1.3B text-to-video model. 
The training time per case is approximately 15 minutes for Content LoRA, 10 minutes for Style LoRA, and 30 minutes for Motion LoRA.

\noindent\textbf{\emph{Dataset.}} We conduct comprehensive evaluations on our proposed benchmark. The dataset comprises 20 content subjects, 32 style references (including 22 artistic styles and 10 material textures), and 20 motion patterns (consisting of 10 object motions and 10 camera movements). To assess the model across the four distinct tasks defined in our benchmark, we curated 200 specific text prompts for each task, representing diverse Content-Style-Motion combinations. In total, we generated and evaluated 800 unique videos to verify the customization fidelity and accuracy of our method.

\begin{figure*}[t]
    \centering
    \includegraphics[width=0.85\textwidth]{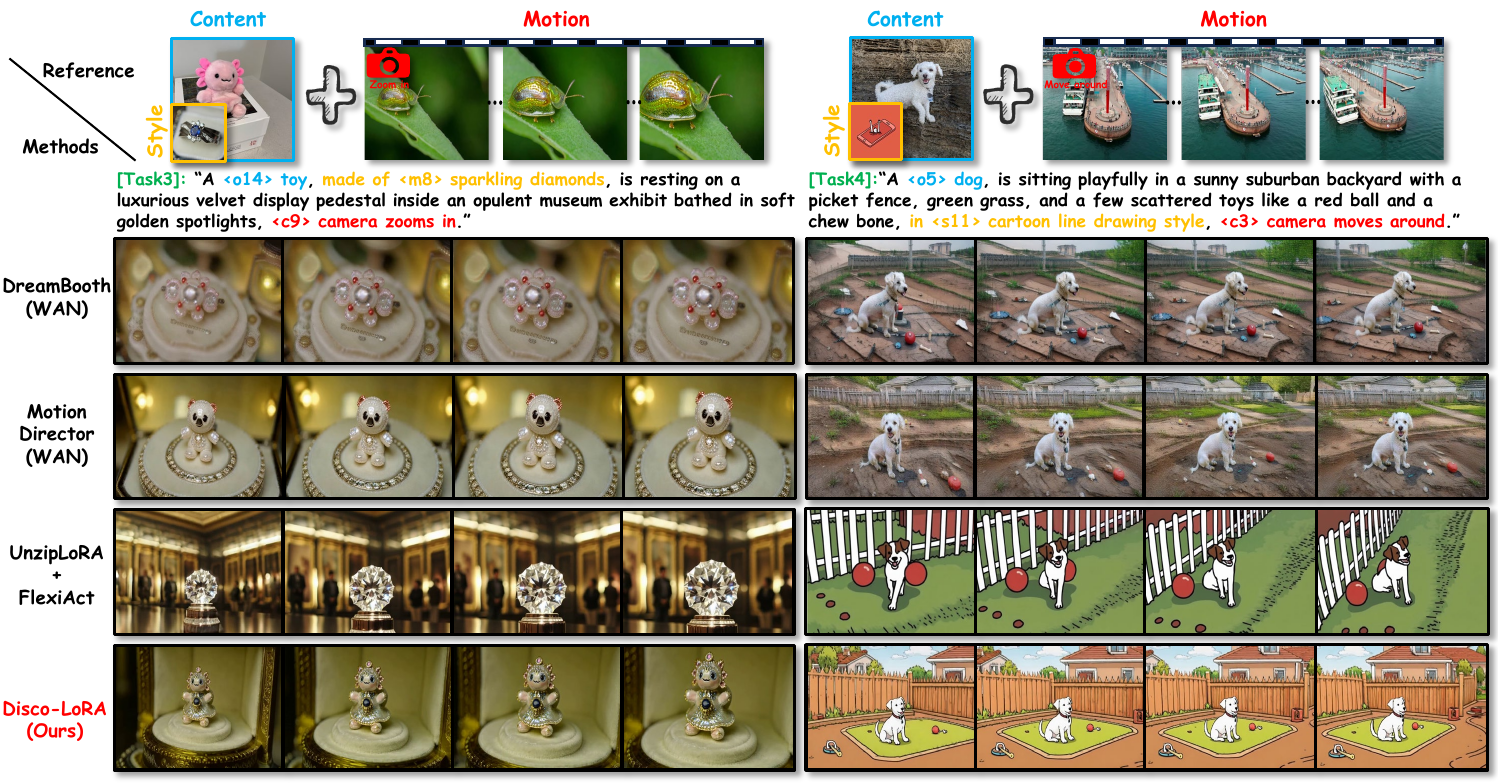}
    \vspace{-10pt}
    \caption{\textbf{Qualitative comparison of Multi-concept Video Customization for Task3 and Task4.} Disco-LoRA preserves content identity, style similarity and camera motion patterns, while other methods fail to stay faithful to the reference.}
    \label{fig_qua2}
    \vspace{-0.2cm}
\end{figure*}
\noindent\textbf{\emph{Metrics.}} Following prior works~\cite{huang2024vbench,zheng2025vbench,chen2025jointtuner,liu2025unziplora,somepalli2024measuring,xu2025clgc}, we evaluate performance using nine metrics across three dimensions:
\noindent(1) Semantic Alignment: We assess text-video consistency via \textbf{CLIP-T} and image-video consistency via \textbf{CLIP-I}~\cite{radford2021learning}. To disentangle style and content, we use three CLIP-I variants: \textbf{CLIP-I(S)} for style reference similarity, \textbf{CLIP-I(C)} for content reference similarity, and \textbf{CLIP-I(A)} for the average of both. Furthermore, to achieve a more robust style evaluation, we employ \textbf{Contrastive Style Descriptors (CSD)}~\cite{somepalli2024measuring} to measure style similarity.
\noindent(2) Motion Quality: We assess dynamics using \textbf{Motion Fidelity} (alignment), \textbf{Subject Consistency} (coherence), and \textbf{Motion Smoothness} (coherence).
\noindent(3) Perceptual Quality: We evaluate visual appeal using \textbf{PickScore} (human preference), \textbf{Aesthetic Quality}, and \textbf{Imaging Quality}.

\renewcommand{\arraystretch}{0.9} 
\begin{table}[t]
\centering
\caption{Quantitative User Studies on Compared Methods.}
\vspace{-10pt}
\scriptsize
\setlength{\tabcolsep}{1.5pt} 
\begin{tabular}{l|c|c|c|c|c}\toprule
    Method
    & \makecell[c]{Prompt\\Alignment}
    & \makecell[c]{Motion\\Similarity}
    & \makecell[c]{Content\\Similarity}
    & \makecell[c]{Style\\Similarity}
    & \makecell[c]{Video\\Quality}
    \\ \midrule
    DreamBooth     & 2.71$\pm$0.16 & 2.09$\pm$0.13 & 4.03$\pm$0.12 & 2.29$\pm$0.14 & 2.56$\pm$0.15 \\
    MotionDirector & 3.17$\pm$0.15 & 3.76$\pm$0.15  & 4.01$\pm$0.15  & 3.32$\pm$0.14 & 3.35$\pm$0.16  \\
    UnzipLoRA+FlexiAct & 3.66$\pm$0.17 & 3.26$\pm$0.14 & 3.39$\pm$0.15 & 3.93$\pm$0.17  & 4.02$\pm$0.16 \\
    \midrule
    Disco-LoRA     & \textbf{4.31$\pm$0.13} & \textbf{4.07$\pm$0.14} & \textbf{4.14$\pm$0.14} & \textbf{4.09$\pm$0.13} & \textbf{4.34$\pm$0.12} \\
    \bottomrule
\end{tabular}
\label{table_userstudy}
\vspace{-0.5cm}
\end{table}
\noindent\textbf{\emph{Compared Methods.}} To evaluate the effectiveness of Disco-LoRA, we compare it with State-of-the-Art (SOTA) approaches on a DiT-based backbone. Since there are limited methods directly comparable on the WAN architecture, we adapt \textbf{DreamBooth}~\cite{ruiz2023dreambooth} and \textbf{MotionDirector}~\cite{zhao2024motiondirector} to the WAN base model for fair comparison. Additionally, we compare against the image customization model \textbf{UnzipLoRA}~\cite{liu2025unziplora} with the motion customization model \textbf{FlexiAct}~\cite{zhang2025flexiact}, utilizing the latter in an Image-to-Video (I2V) setting to achieve motion customization.

\subsection{Qualitative Evaluation}
We evaluate Disco-LoRA against DreamBooth~\cite{ruiz2023dreambooth}, MotionDirector~\cite{zhao2024motiondirector}, and UnzipLoRA~\cite{liu2025unziplora}+FlexiAct~\cite{zhang2025flexiact}. Figures~\ref{fig_qua1} and \ref{fig_qua2} demonstrate our method's superior capability in disentangling content, style, and motion compared with these baselines.

In Fig.~\ref{fig_qua1}, baseline methods exhibit significant limitations in consistency. DreamBooth and MotionDirector struggle to maintain style alignment throughout the video sequence. Furthermore, UnzipLoRA+FlexiAct fails to generate a coherent initial frame that respects the target style, which cascades into motion customization failures. A clear example is the "playing piano" case, where the subject fails to interact physically with the environment. In contrast, Disco-LoRA generates videos that preserve fine-grained content details while maintaining artistic styles and material textures consistent with the reference images. It accurately reproduces target motion patterns, demonstrating superior disentanglement performance compared to prior methods.

In Fig.~\ref{fig_qua2}, DreamBooth suffers from severe content entanglement by erroneously blending the appearance of the jewelry with the toy, while standard LoRA training proves insufficient for learning precise camera motion. Although MotionDirector captures the camera trajectory, its integration with style transfer compromises content fidelity, leading to geometric hallucinations and inconsistent local style application. UnzipLoRA+FlexiAct appears heavily biased by the text prompt, misinterpreting the content as a diamond and introducing semantic errors during style composition, such as rendering the dog with incorrect ear colors. Conversely, Disco-LoRA avoids these artifacts, generating videos that faithfully retain content and style details while effectively reproducing complex motion patterns.

\subsection{Quantitative Evaluation}
To evaluate the visual quality of synthesized videos, we conduct a comprehensive comparison between Disco-LoRA and state-of-the-art approaches, including DreamBooth~\cite{ruiz2023dreambooth}, MotionDirector~\cite{zhao2024motiondirector}, and the combined pipeline of UnzipLoRA~\cite{liu2025unziplora} and FlexiAct~\cite{zhang2025flexiact}. As illustrated in Tab.~\ref{table_comparison} and ~\ref{table_userstudy}, Disco-LoRA demonstrates robust capabilities in customized generation, successfully disentangling and controlling content, style, and motion across all tasks.

\noindent\textbf{\emph{Objective Evaluation.}}  
As shown in Tab.~\ref{table_comparison}, Disco-LoRA demonstrates superior performance across semantic, motion, and perceptual metrics:
\noindent(1) \textbf{Alignment and Consistency:} Disco-LoRA achieves the highest text-video alignment (CLIP-T: \textbf{0.361}) and leads in style-related metrics (CLIP-I (S): \textbf{0.592}, CSD: \textbf{0.211}, CLIP-I (A): \textbf{0.633}). While baselines like MotionDirector show higher CLIP-I (C), this reflects overfitting to the subject at the expense of style fidelity.
\noindent(2) \textbf{Motion Quality:} Our method excels in motion generation with a Motion Fidelity of \textbf{0.771}, significantly outperforming UnzipLoRA+FlexiAct (0.491) and MotionDirector (0.701). Simultaneously, it maintains high Subject Consistency (0.971) and exceptional Motion Smoothness (\textbf{0.987}).
\noindent(3) \textbf{Perceptual Quality:} Disco-LoRA dominates quality-centric metrics, securing top scores in PickScore (\textbf{0.212}), Aesthetic Quality (\textbf{0.683}), and Imaging Quality (\textbf{0.694}), validating its ability to generate visually superior videos

\noindent\textbf{\emph{User Study.}}  
We conducted user studies to comprehensively evaluate the effectiveness of Disco-LoRA. A total of 100 participants voted on 80 pairs of customized videos for four tasks, which were randomly presented to reduce bias. The evaluation was conducted based on five criteria: prompt alignment, motion similarity, appearance similarity, style similarity, and video quality. Each video was rated on a scale of 1 to 5, resulting in 8,000 ratings.
As shown in Tab.~\ref{table_userstudy}, Disco-LoRA achieved the highest scores across all five criteria compared to other SOTA methods. Notably, our method scored above 4.0 in every category, with particularly strong performance in Prompt Alignment and Video Quality.
We also performed an analysis on the $95\%$ confidence interval and found that Disco-LoRA outperformed baseline methods with statistically significant results, demonstrating its robustness in generating high-quality, stylistically consistent, and motion-faithful videos.

\subsection{Ablation Study}
To assess the contribution of each component in Disco-LoRA, we conduct comprehensive ablation studies focusing on its core modules, as summarized in Table~\ref{table_comparison}.

\noindent\textbf{Effect of Complementary Prompting Strategy. (C.P)}
The complementary prompting strategy is crucial for promoting the learning of each concept. Without this constraint, LoRA cannot learn its respective concepts effectively. Consequently, the Target LoRA fails to learn the specific desired concept, leading to inconsistent generation results. This degradation is reflected in lower scores across CLIP-I(S), CSD, CLIP-I(C), and CLIP-I(A).

\noindent\textbf{Effect of Time-aware Masking Strategy. (T.A.M)} 
The time-aware masking strategy effectively distinguishes and constrains the disentanglement of different concepts. In the absence of this strategy, the Style LoRA tends to overfit the appearance information of the reference artistic image (e.g., learning the texture along with the underlying object structure). While this may result in a higher CLIP-I(S) score, it stems from overfitting to appearance features rather than true style transfer, ultimately causing a decrease in content and action fidelity, as evidenced by lower CLIP-I(C) and CLIP-I(A).

\noindent\textbf{Effect of $L_{trend}$.}
The $L_{trend}$ loss is designed to constrain the optimization trajectory of each LoRA to align with the original trend. Removing this constraint prevents the model from accurately capturing the individual concepts, resulting in poor similarity performance across all metrics: CLIP-I(S), CSD, CLIP-I(C), and CLIP-I(A).

\noindent\textbf{Effect of $L_{mag}$.}
The purpose of $L_{mag}$ is to regularize the magnitude of different LoRAs, ensuring their means remain consistent so that every provided concept is adequately represented in the video. Without the $L_{mag}$ constraint, the Content LoRA tends to dominate the generation process. While this results in high CLIP-I(C), it suppresses the style transfer, leading to significantly lower CLIP-I(S) and CSD.

\begin{figure}[t]
    \centering
    \includegraphics[width=0.45\textwidth]{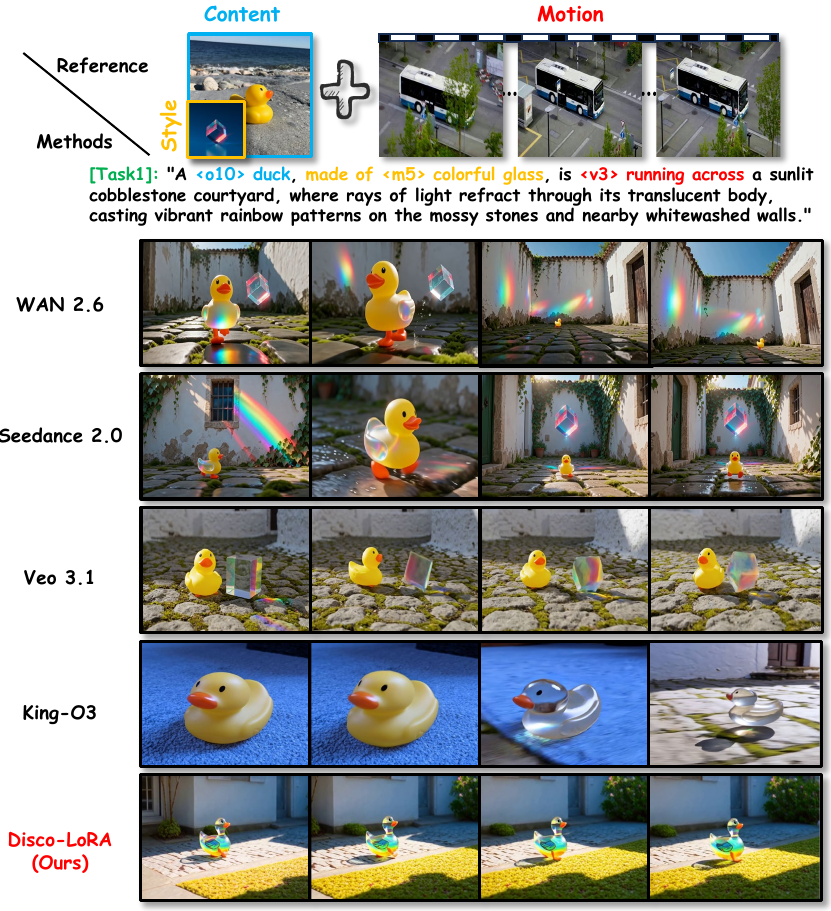}
    \vspace{-10pt}
    \caption{\textbf{Qualitative comparison of multi-concept video customization against commercial methods.} Existing commercial models struggle to simultaneously customize specific subjects, styles, and motions.}
    \label{fig_qua3}
    \vspace{-0.5cm}
\end{figure}
\subsection{Comparison with Commercial Methods}
To demonstrate the limitations of existing commercial models in jointly customizing subject, style, and motion, we compare our approach against SOTA commercial models, including WAN 2.6~\cite{wan2025wan}, Seedance 2.0~\cite{gao2025seedance,seedance2025seedance}, Veo 3.1~\cite{Veo3}, and Kling-O3~\cite{team2025kling}. As illustrated in Fig.~\ref{fig_qua3}, these models fail to effectively integrate the desired style with the subject, and they struggle to adhere to specific motion constraints. In contrast, our proposed Disco-LoRA successfully achieves multi-concept customization, enabling highly controllable generation that extends beyond mere subject preservation to encompass additional conditions.
\section{Conclusion}\label{con}  
In this work, we systematically define the multi-concept video customization task and establish a comprehensive benchmark for future research.
We present Disco-LoRA, a novel framework that enables the disentangled learning and flexible recombination of content, style, and motion. 
By leveraging an Iterative Dual-LoRA Disentanglement Framework for concept separation and a Z-score-based statistical regularization for LoRA weight distribution alignment, our method successfully mitigates interference between heterogeneous modalities during composition.
Extensive experiments demonstrate that Disco-LoRA maintains the base model's generative flexibility while consistently delivering high-quality, text-aligned outputs. Future work will focus on scaling Disco-LoRA to support a larger number of concurrent concepts.
{
    \small
    \bibliographystyle{ieeenat_fullname}
    \bibliography{main}
}

\clearpage
\setcounter{page}{1}
\maketitlesupplementary

\section{Overall}  
The supplementary material includes the following sections:  
\begin{itemize}  
    \item \textcolor{black}{Details of our methods and experiments.}  
    \item \textcolor{black}{Limitations, Discussion and Future Work of our method.}  
    \item \textcolor{black}{A demo video introducing our work comprehensively.}  
    \item \textcolor{black}{A folder containing some videos generated by our model.}  
    \item \textcolor{black}{A folder containing the subset of our benchmark.}  
\end{itemize}
\section{Details of our methods and experiments}
\noindent\subsection{Hyperparameters}
For the Iterative Dual-LoRA Disentanglement Framework, we address two sub-tasks. In Content-Style Disentanglement, we train the content LoRA ($\Delta W_c$) for 3 epochs with $T_l=0.5T$ and $T_h=T$, and the style LoRA ($\Delta W_s$) for 4 epochs with $T_l=0.4T$ and $T_h=0.8T$. For Content-Motion Disentanglement, the motion LoRA ($\Delta W_m$) is trained for 5 epochs with $T_l=0.5T$ and $T_h=T$.
Regarding the Z-Score-Based Statistical Regularization, We set the regularization hyperparameters to $\lambda_{mag}=1.0$ and $\lambda_{trend}=0.05$ to balance the magnitudes of different loss components. We also introduce a scaling factor of $s=10.0$ to match the numerical range.
\noindent\subsection{User Study}\label{sub:USI}
\begin{figure}[t]
    \centering
    \includegraphics[width=0.45\textwidth]{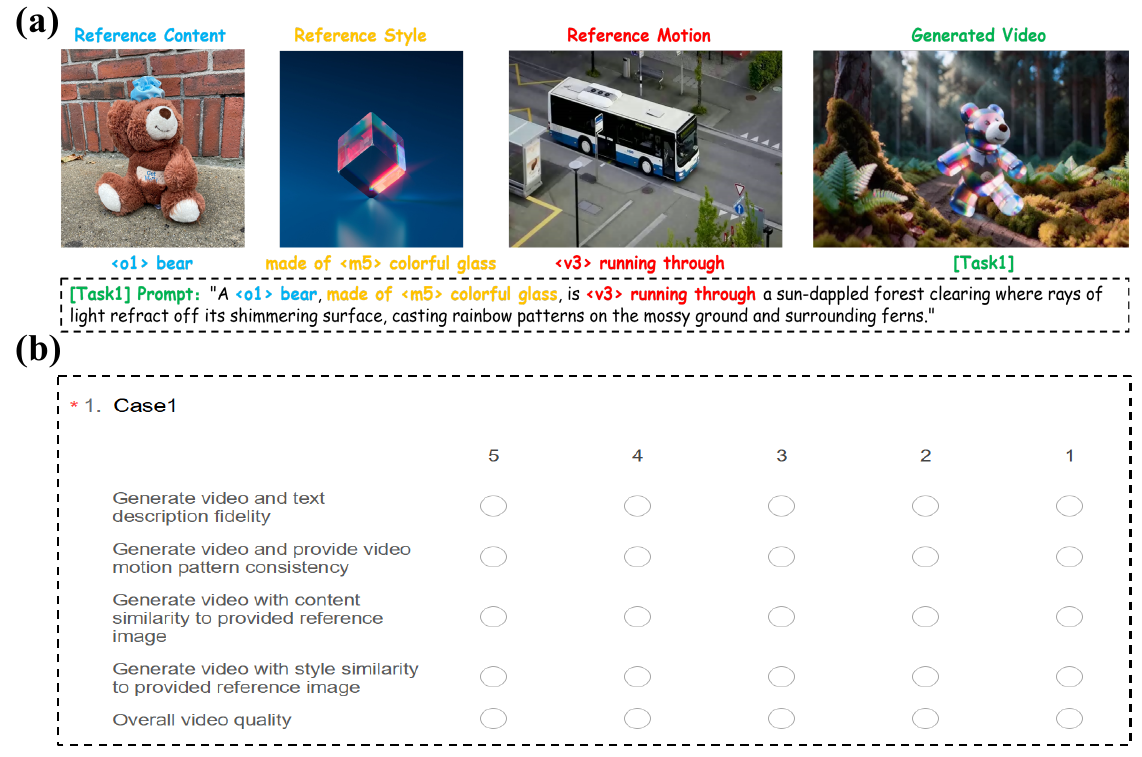}
    \vspace{-10pt}
    \caption{\textbf{Human evaluation questionnaire format.} (a) presents the reference subject image, reference video, and the customized video generated by the model. Participants are then asked to complete the evaluation form in (b), rating the quality of the generated video based on Prompt Alignment, Motion Similarity, Appearance Similarity and Video Quality.}
    \label{fig_userstudy}
    \vspace{-0.2cm}
\end{figure}

During the user study, we provide each case video generated by DreamBooth(WAN), MotionDirector(WAN), UnzipLoRA+FlexiAct, and our Disco-LoRA for evaluation based on five questions. Each question is rated on a scale from 1 to 5 for the following criteria:  
(1) The accuracy of generating the video to match the text descriptions (\textit{Prompt Alignment}).  
(2) Consistency between the generated video and the provided motion mode (\textit{Motion Similarity}).  
(3) The similarity between the main content of the generated video and the reference image provided (\textit{Content Similarity}).  
(4) The similarity between the main style/material of the generated video and the reference image provided (\textit{Style Similarity}).
(5) The overall quality of the video (\textit{Video Quality}).
Fig.~\ref{fig_userstudy} shows the format of our questionnaire.

\begin{figure}[t]
    \centering
    \includegraphics[width=0.45\textwidth]{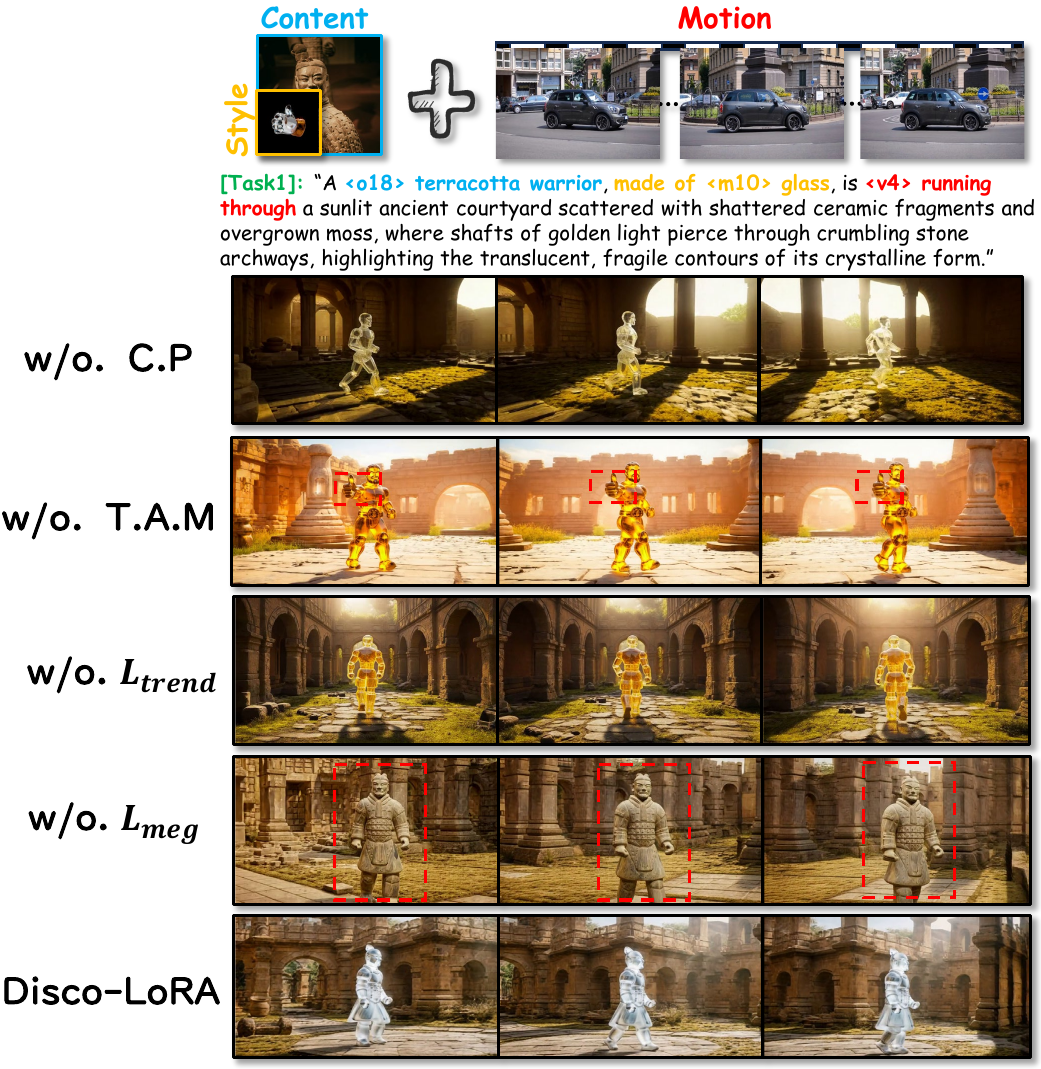}
    \vspace{-10pt}
    \caption{Qualitative Evaluation For Ablation Study.}
    \label{fig_ablation}
    \vspace{-0.5cm}
\end{figure}

\noindent\subsection{Qualitative Evaluation For Ablation Study}
To assess the contribution of each component within Disco-LoRA, we conducted comprehensive ablation studies focusing on its core modules, as summarized in Fig.~\ref{fig_ablation}. While the quantitative results were discussed in the main text, here we evaluate the effectiveness of each module from a qualitative perspective.

\noindent\textbf{Effect of Complementary Prompting Strategy.(C.P)}
The complementary prompting strategy is crucial for promoting the learning of each concept. In the absence of this constraint, the disentanglement LoRA cannot learn its respective concepts effectively. As a result, generation degenerates to relying exclusively on the WAN's native semantic priors.

\noindent\textbf{Effect of Time-aware Masking Strategy.(T.A.M)} 
The time-aware masking strategy effectively distinguishes and constrains the disentanglement of different concepts. In the absence of this strategy, the Style LoRA tends to overfit the appearance information of the reference artistic image. For instance, the anomalous appearance of a 'thumb-up' gesture results from the model overfitting to the global shape of the material during style learning, leading to structurally implausible generations. 

\noindent\textbf{Effect of $L_{trend}$.}
The $L_{trend}$ loss is designed to constrain the optimization trajectory of each LoRA, ensuring adherence to the original trend. Removing this constraint hinders the model's ability to accurately capture individual concepts, leading to suboptimal similarity performance across all provided reference concepts.

\noindent\textbf{Effect of $L_{mag}$.}
The purpose of $L_{mag}$ is to regularize the magnitude of different LoRAs, ensuring their means remain consistent so that every provided concept is adequately represented in the final video. Without the $L_{mag}$ constraint, the Content LoRA dominates the generation process; for instance, the Terracotta Warriors in the figure fail to exhibit any of the intended stylistic attributes.
\noindent\subsection{Iterative Dual-LoRA Disentanglement}

\begin{figure}[t]
    \centering
    \includegraphics[width=0.4\textwidth]{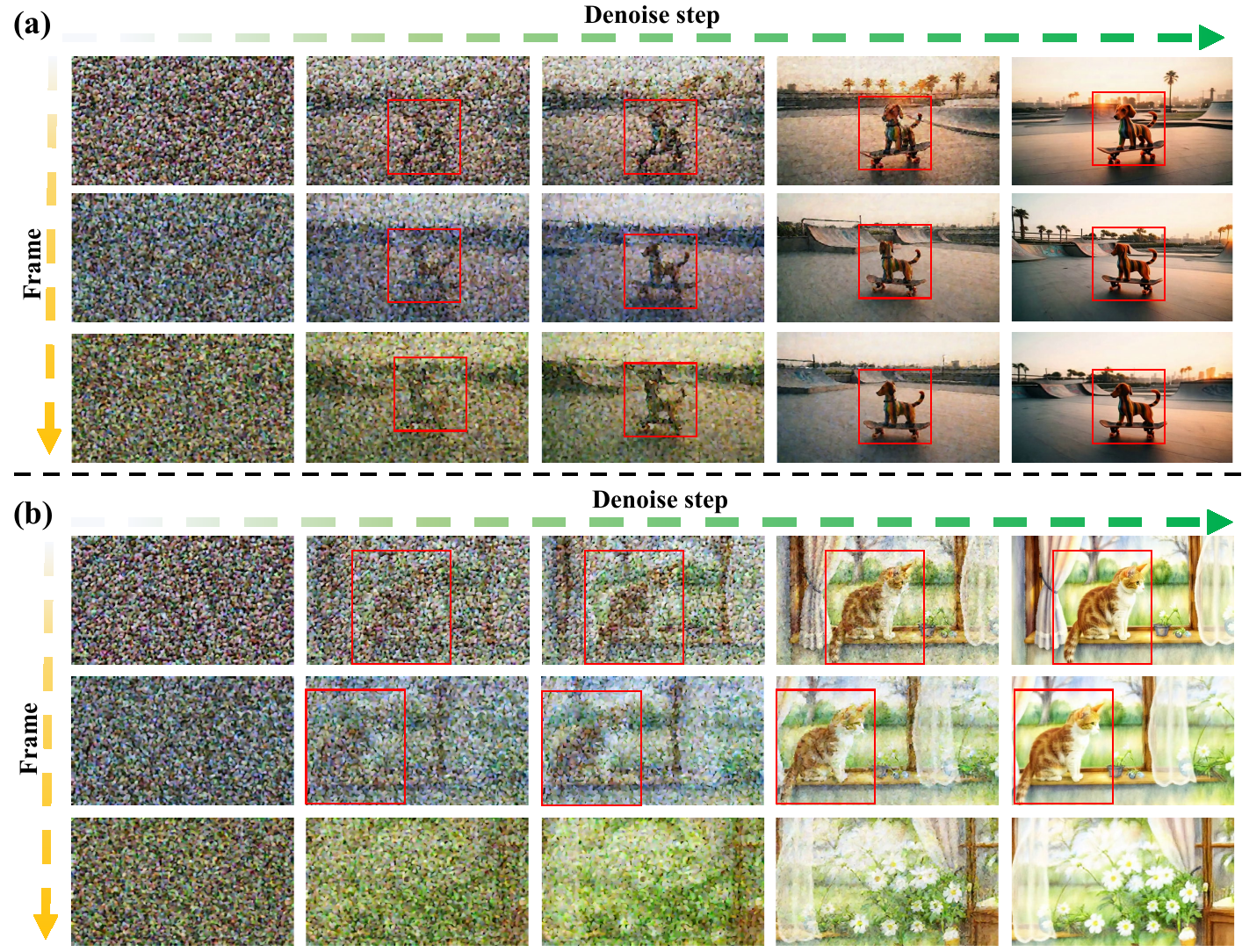}
    \vspace{-10pt}
    \caption{Visualization of the denoising timesteps for Style. (a) The denoising process for material styles. In the early stages of diffusion, the model focuses on generating low-frequency information such as object position and contours, while in the later stages, it synthesizes specific object textures. (b) The denoising process for artistic styles. Similarly, the model prioritizes structural elements in the early stages, but shifts to generating global stylistic details in the final steps.}
    \label{fig_timestep}
\end{figure}

\begin{figure}[t]
    \centering
    \includegraphics[width=0.4\textwidth]{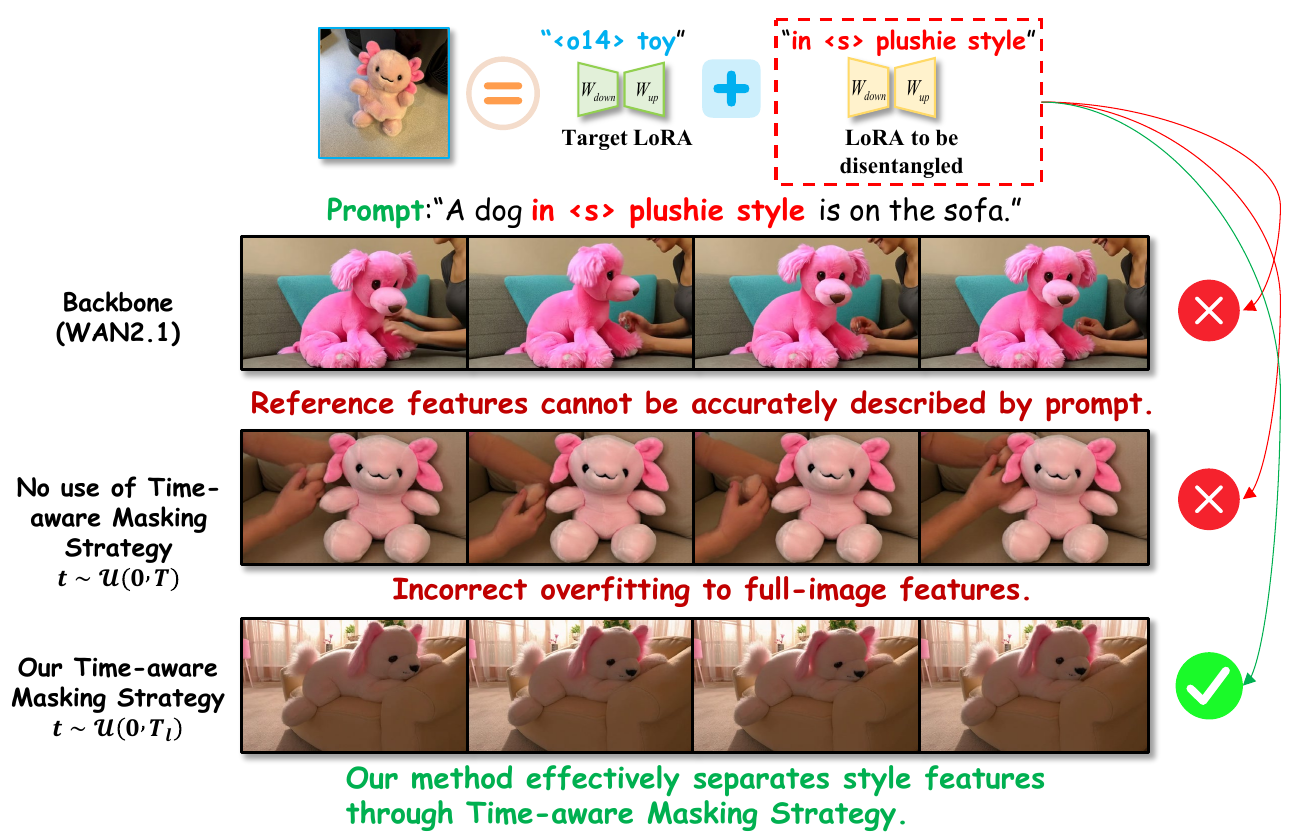}
    \vspace{-10pt}
    \caption{Ablation study on the Time-aware Masking Strategy. This study demonstrates that our strategy enables the two LoRAs to focus on their respective text-aligned features, successfully achieving disentanglement. Without distinguishing between timesteps, the auxiliary LoRA overfits to the entire frame, rendering the target LoRA ineffective during inference.}
    \label{fig_timestep2}
\end{figure}

\begin{algorithm}[htb]
\caption{Iterative Dual-LoRA Disentanglement for \textbf{\textcolor{red}{$\Delta W_s$}}}
\label{alg:iterative_dual_lora_ws}
\begin{algorithmic}[1]
\Require Total epochs $\mathcal{N}$
\Require Pre-trained weights $W_0$, training data $\mathcal{D}$
\Require Style LoRA $\Delta W_s$ (Target LoRA) and  Content LoRA $\Delta W_c$ (LoRA to be disentangled)
\Require Prompts $P_s$ (e.g., ``A $\langle$content$\rangle$, in $\langle$style$\rangle$'')
\Require Prompts $P_c$ (e.g., ``A $\langle$content$\rangle$'')

\Require Timestep thresholds $T_l, T_h$ and total timesteps $T$, where $0 \leq T_l \leq T_h \leq T$

\State \textbf{for} each training epoch $i \in \mathcal{N}$ \textbf{do}
\Statex \hspace{\algorithmicindent} \textcolor{gray}{\textbf{//Phase 1: Update Content LoRA}}
\State \hspace{\algorithmicindent} \textbf{for} each training step or batch $x \in \mathcal{D}$ \textbf{do}
    \State \hspace{\algorithmicindent}\hspace{1em} \textcolor{blue}{\textbf{Freeze} $\Delta W_s$}; \textcolor{red}{\textbf{Update} $\Delta W_c$}
    \State \hspace{\algorithmicindent}\hspace{1em} Sample timestep $t \sim \mathcal{U}(T_l, T_h)$
    \State \hspace{\algorithmicindent}\hspace{1em} $W \leftarrow W_0 + \Delta W_c + \text{StopGrad}(\Delta W_s)$
    \State \hspace{\algorithmicindent}\hspace{1em} $\mathcal{L}_c \leftarrow \mathcal{L}_{\text{diff}}(x, P_c, t; W)$
    \State \hspace{\algorithmicindent}\hspace{1em} $\Delta W_c \leftarrow \Delta W_c - \eta \frac{\partial \mathcal{L}_c}{\partial \Delta W_c}$
\State \hspace{\algorithmicindent} \textbf{end for}

\Statex \hspace{\algorithmicindent} \textcolor{gray}{\textbf{//Phase 2: Update Style LoRA}}
\State \hspace{\algorithmicindent} \textbf{for} each training step or batch $x \in \mathcal{D}$ \textbf{do}
    \State \hspace{\algorithmicindent}\hspace{1em} \textcolor{red}{\textbf{Update} $\Delta W_s$}; \textcolor{blue}{\textbf{Freeze} $\Delta W_c$}
    \State \hspace{\algorithmicindent}\hspace{1em} Sample timestep $t \sim \mathcal{U}(0, T_l)$
    \State \hspace{\algorithmicindent}\hspace{1em} $W \leftarrow W_0 + \text{StopGrad}(\Delta W_c) + \Delta W_s$
    \State \hspace{\algorithmicindent}\hspace{1em} $\mathcal{L}_s \leftarrow \mathcal{L}_{\text{diff}}(x, P_s, t; W)$
    \State \hspace{\algorithmicindent}\hspace{1em} $\Delta W_s \leftarrow \Delta W_s - \eta \frac{\partial \mathcal{L}_s}{\partial \Delta W_s}$
\State \hspace{\algorithmicindent} \textbf{end for}
\State \textbf{end for}
\State \textbf{Return} \textbf{\textcolor{red}{$\Delta W_s$}}
\end{algorithmic}
\end{algorithm}
\begin{algorithm}[htb]
\caption{Iterative Dual-LoRA Disentanglement for \textbf{\textcolor{red}{$\Delta W_m$}}}
\label{alg:iterative_dual_lora_wm}
\begin{algorithmic}[1]
\Require Total epochs $\mathcal{N}$
\Require Pre-trained weights $W_0$, training data $\mathcal{D}$
\Require Motion LoRA $\Delta W_m$ (Target LoRA) and  Content LoRA $\Delta W_c$ (LoRA to be disentangled)
\Require Prompts $P_m$ (e.g., ``$\langle$content$\rangle$,  $\langle$motion$\rangle$'')
\Require Prompts $P_c$ (e.g., ``$\langle$content$\rangle$'')

\Require Timestep thresholds $t_l, t_h$ and total timesteps $T$, where $0 \leq T_l \leq T_h \leq T$

\State \textbf{for} each training epoch $i \in \mathcal{N}$ \textbf{do}
\Statex \hspace{\algorithmicindent} \textcolor{gray}{\textbf{//Phase 1: Update Content LoRA}}
\State \hspace{\algorithmicindent} \textbf{for} each training batch $X \in \mathcal{D}$ \textbf{do}
    \State \hspace{\algorithmicindent}\hspace{1em} \textcolor{blue}{\textbf{Freeze} $\Delta W_m$}; \textcolor{red}{\textbf{Update} $\Delta W_c$}
    \State \hspace{\algorithmicindent}\hspace{1em} $x \leftarrow X[0]$ //Get the first frame
    \State \hspace{\algorithmicindent}\hspace{1em} Sample timestep $t \sim \mathcal{U}(T_l, T_h))$
    \State \hspace{\algorithmicindent}\hspace{1em} $W \leftarrow W_0 + \Delta W_c$
    \State \hspace{\algorithmicindent}\hspace{1em} $\mathcal{L}_c \leftarrow \mathcal{L}_{\text{diff}}(x, P_c, t; W)$
    \State \hspace{\algorithmicindent}\hspace{1em} $\Delta W_c \leftarrow \Delta W_c - \eta \frac{\partial \mathcal{L}_c}{\partial \Delta W_c}$
\State \hspace{\algorithmicindent} \textbf{end for}

\Statex \hspace{\algorithmicindent} \textcolor{gray}{\textbf{//Phase 2: Update Motion LoRA}}
\State \hspace{\algorithmicindent} \textbf{for} each training step or batch $X \in \mathcal{D}$ \textbf{do}
    \State \hspace{\algorithmicindent}\hspace{1em} \textcolor{red}{\textbf{Update} $\Delta W_c$}; \textcolor{blue}{\textbf{Freeze} $\Delta W_m$}
    \State \hspace{\algorithmicindent}\hspace{1em} Sample timestep $t \sim \mathcal{U}(0,T_l)$
    \State \hspace{\algorithmicindent}\hspace{1em} $W \leftarrow W_0 + \text{StopGrad}(\Delta W_c) + \Delta W_m$
    \State \hspace{\algorithmicindent}\hspace{1em} $\mathcal{L}_m \leftarrow \mathcal{L}_{\text{diff}}(X, P_m, t; W)$
    \State \hspace{\algorithmicindent}\hspace{1em} $\Delta W_m \leftarrow \Delta W_m - \eta \frac{\partial \mathcal{L}_m}{\partial \Delta W_m}$
\State \hspace{\algorithmicindent} \textbf{end for}
\State \textbf{end for}
\State \textbf{Return} \textbf{\textcolor{red}{$\Delta W_m$}}
\end{algorithmic}
\end{algorithm}
While Sec. 3.3 primarily focuses on the training of $\Delta W_c$ with its algorithm, our framework also involves learning the style ($\Delta W_s$) and motion ($\Delta W_m$) components. To provide a comprehensive view of the training process, we present the algorithms for these additional components here: Algorithm~\ref{alg:iterative_dual_lora_ws} details the optimization of $\Delta W_s$, and Algorithm~\ref{alg:iterative_dual_lora_wm} outlines the procedure for $\Delta W_m$.

In Sec.~3.3, we introduce the Time-aware Masking Strategy. By leveraging the inherent denoising properties of diffusion models, we explicitly distinguish the learning of different concepts across varying timesteps \(t \sim \mathcal{U}(0, T)\). 
Specifically, we target motion during the early denoising stage (high noise levels, large \(t\)), where the model focuses on establishing object positioning and overall scene layout (Fig.~\ref{fig_timestep}). Subsequently, we capture content during the intermediate stage (medium noise levels, moderate \(t\)), as objects begin to exhibit distinct contours and specific morphological characteristics. Finally, we focus on style in the late stage (low noise levels, small \(t\)). Once the object's shape and appearance are largely determined, the model shifts its attention to fine-grained details such as color tones and textures. This progression remains consistent across different domains, including materials (Fig.~\ref{fig_timestep}(a)) and artistic styles (Fig.~\ref{fig_timestep}(b)).

Furthermore, as shown in Fig.~\ref{fig_timestep2}, our Time-aware Masking Strategy enables the two LoRAs to focus on their respective text-aligned features. Without distinguishing between timesteps, the disentangling LoRA overfits to the entire frame, rendering the target LoRA ineffective during inference. Our selection of timesteps leverages the findings from~\cite{zhang2023prospect}, effectively disentangling different concepts within the same image.

\noindent\subsection{Benchmark build pipeline}
\begin{figure*}[t]
    \centering
    \includegraphics[width=0.95\textwidth]{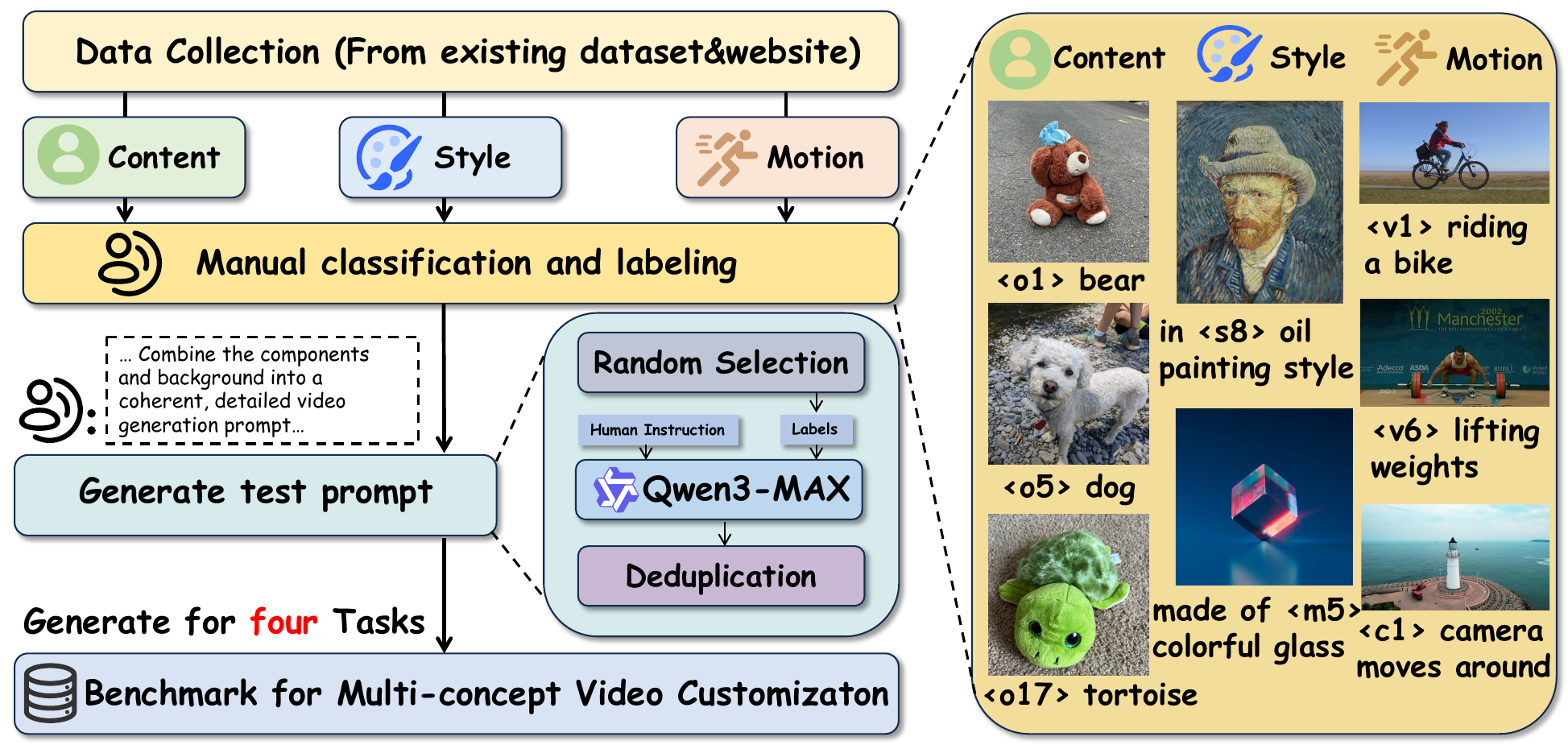}
    \vspace{-10pt}
    \caption{\textbf{Construction pipeline of our benchmark.}}
    \label{fig_dataset}
\end{figure*}

In Sec. 3.2, we briefly introduced the data sources and the construction process of our benchmark. In this section, we provide a detailed elaboration of this pipeline, as illustrated in Fig.~\ref{fig_dataset}.
First, we collected 20 distinct content images from the existing image personalization benchmark, DreamBench~\cite{ruiz2023dreambooth}. For style references, we gathered 22 categories of distinct artistic style images from StyleDrop~\cite{sohn2023styledrop}, and 10 images with significant material textures from internet searches and Prospect~\cite{zhang2023prospect}.
Furthermore, we curated videos featuring distinct object motion from the Davis~\cite{darcet2023vitneedreg} dataset. Obtaining data with fixed camera movements proved challenging, as camera motion is often coupled with subject movement. To address this, we collected aerial footage of static objects from the internet, which meets our requirements. Additionally, we sourced I2V camera motion results from the open-source VAP~\cite{bian2025video} dataset. Since the raw quality of the VAP dataset was suboptimal, we manually filtered the data to ensure usability.

Subsequently, we manually categorized the collected data following the taxonomy defined in Sec. 3.2: dividing Style into Material Style and Artistic Style, and Motion into Object Motion and Camera Movement. For each data sample, we manually assigned a specific label to represent the target concept and drafted a detailed descriptive caption to supervise the reconstruction loss during training.

Finally, to generate test prompts for the four tasks, we employed the Qwen3-MAX~\cite{yang2025qwen3} model. Specifically, each task requires combining three distinct concepts. To ensure flexibility and diversity in these combinations, we randomly sampled three concept categories from our labels for each instance. We guided Qwen3-MAX to generate the test prompts using the following instruction:
\begin{tcolorbox}[colback=gray!10, colframe=black, arc=0mm, boxrule=1pt]
\small
You are a creative prompt generator. Your task is to: 
1. I will provide you with three elements: the object, the material, and the motion. For example: \texttt{<o1> plushie bear}, \texttt{<m1> soft fabric}, \texttt{<v1> running}. I will combine these three elements into a complete JSON-formatted prompt. 
2. Generate a suitable background that complements these components. 
3. Combine the components and background into a coherent, detailed image generation prompt. The specific format should be: A \texttt{<object>}, made of \texttt{<material>}, is \texttt{<motion>} \texttt{<background>}. 

The prompt should:
\begin{itemize}
    \item Include all the provided components naturally placed in the scene
    \item Describe a realistic and harmonious background
    \item Use natural English language
    \item Keep the component identifiers (like \texttt{<i1>}, \texttt{<s1>}, \texttt{<m1>}, \texttt{<o1>}, \texttt{<v1>}, \texttt{<c1>}) in the final prompt.
\end{itemize}

Please generate a complete image generation prompt based on the following JSON input. Generate a background that fits these components and combine everything into one cohesive prompt following the format: A \texttt{<object>}, made of \texttt{<material>}, is \texttt{<motion>} \texttt{<background>}
\end{tcolorbox}
\noindent\subsection{Details for Metrics}
We establish a comprehensive evaluation framework across three dimensions: Semantic Alignment, Motion Quality and Perceptual Quality, using nine metrics.
\begin{itemize}
\item \textcolor{black}{\textbf{Semantic Alignment.}} 
(1) \textbf{CLIP-T:} This metric evaluates the alignment between text prompts and generated videos by calculating the average frame-wise cosine similarity between their embeddings extracted via the CLIP~\cite{radford2021learning} model.
(2) \textbf{CLIP-I(S):} This metric quantifies the visual similarity between the reference style images and the generated video frames. It computes the cosine similarity of their embeddings using the CLIP image encoder~\cite{radford2021learning}.
(3) \textbf{CLIP-I(C):} Similar to CLIP-I(S), this metric measures the visual correspondence between the reference content images and the generated video frames based on their CLIP image embeddings.
(4) \textbf{CLIP-I(A):} This metric represents the overall performance by calculating the arithmetic mean of CLIP-I(S) and CLIP-I(C).
(5) \textbf{CSD:} This metric~\cite{somepalli2024measuring} evaluates the pure stylistic alignment between the reference style images and the generated video frames using Contrastive Style Descriptors. Unlike CLIP-based metrics that may entangle style with semantic content, CSD extracts representations using a Vision Transformer (ViT) backbone trained via multi-label contrastive learning to specifically capture style-related visual attributes (e.g., colors, textures, and brushstrokes) while remaining invariant to the underlying semantic content. The metric computes the cosine similarity (or dot product) between the embeddings of the generated frames and the reference images.
\item \textcolor{black}{\textbf{Motion Quality.}}  
(1) \textbf{Motion Fidelity:} Evaluates the consistency of motion patterns by leveraging CoTracker3~\cite{karaev2024cotracker}, a model designed for diffusion-motion-transfer~\cite{yatim2024space}.  
(2) \textbf{Subject Consistency:} Assesses whether the appearance of the subject (e.g., characters) remains consistent across different frames in the video, as implemented in VBench~\cite{huang2024vbench}.  
(3) \textbf{Motion Smoothness:} Evaluates the temporal coherence of the generated videos by quantifying frame-to-frame motion consistency, following the implementation in VBench~\cite{huang2024vbench}.
\item \textcolor{black}{\textbf{Perceptual Quality.}}  
(1) \textbf{PickScore:} Predicts human preference scores using PickScore~\cite{kirstain2023pick}, with results averaged at the frame level.  
(2) \textbf{Aesthetic Quality:} Measures artistic merit using the LAION aesthetic predictor, implemented via VBench~\cite{huang2024vbench}.  
(3) \textbf{Imaging Quality:} Evaluates distortions in generated frames, such as overexposure, noise, and blurriness, as assessed via VBench~\cite{huang2024vbench}.
\end{itemize}

\section{Limitations, Discussion and Future Work}

\noindent\textbf{Limitations.} While Disco-LoRA demonstrates exceptional performance in disentangling and customizing content, style, and motion for videos, its current exploration is primarily confined to a specific triad of concepts. A limitation lies in the scalability of concept composition. We have not yet extensively investigated the framework's efficacy when scaling to a higher number of simultaneous concepts (e.g., combining 4 or 5 distinct elements). Furthermore, our current scope focuses heavily on the primary subject and its dynamics, without fully exploring the disentanglement of other critical visual dimensions, such as complex background environments or specific lighting conditions. This restricts the framework's potential for holistic scene construction.
\begin{figure}[t]
    \centering
    \includegraphics[width=0.45\textwidth]{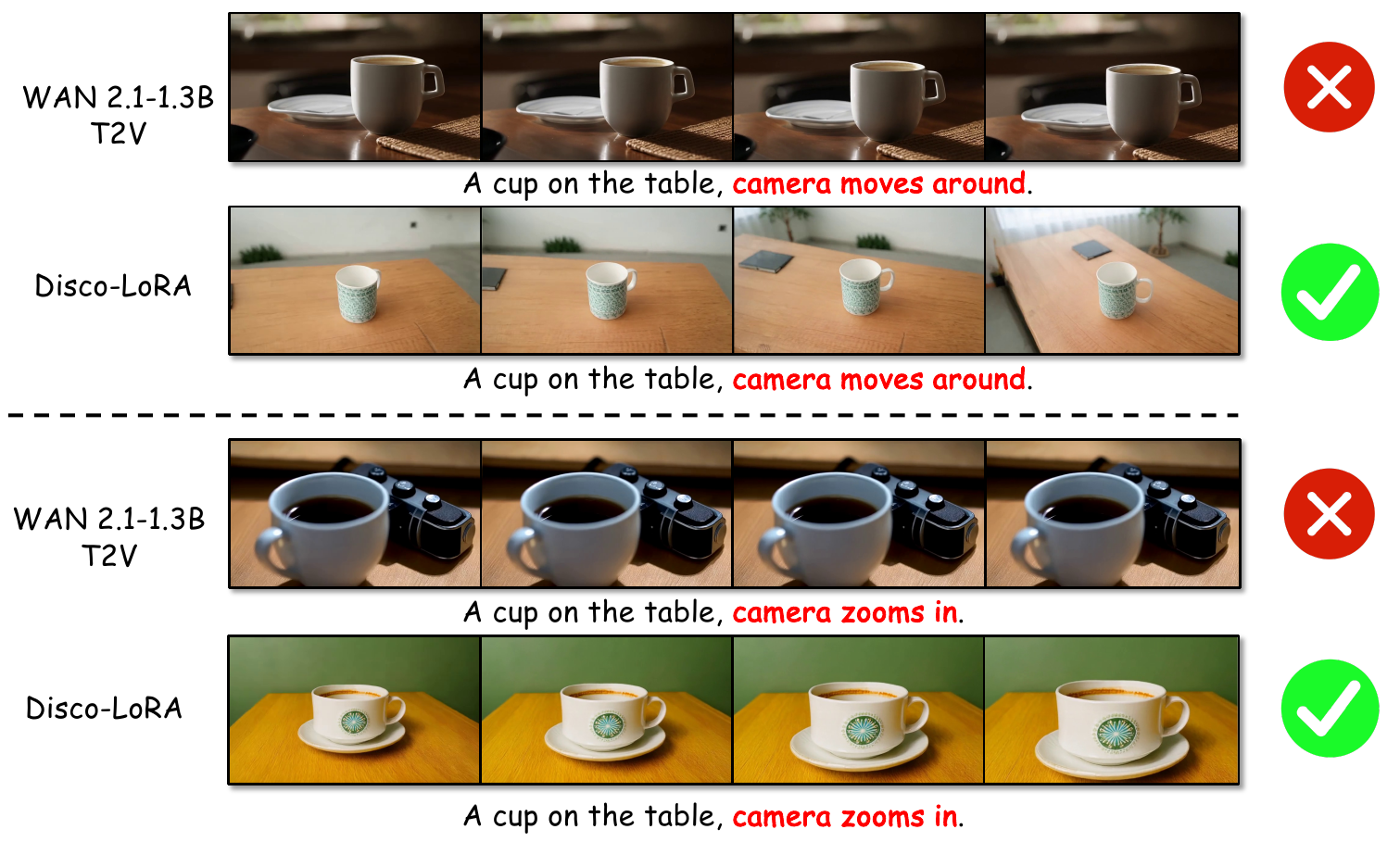}
    \vspace{-10pt}
    \caption{Comparison with backbone with camera control.}
    \label{fig_discussion_camera}
\end{figure}

\noindent\textbf{Discussion.} 
(1) As illustrated in Fig.~\ref{fig_discussion_camera}, we observe that the base model WAN2.1-1.3B-T2V lacks the capability for precise text-driven camera motion control. However, by integrating our trained Motion LoRA, we successfully enable accurate text-guided camera control on this base model. (2) Although methods like VideoMage~\cite{huang2025videomage} achieve multi-subject customization, they report an inability to extend this capability to non-subject concepts (e.g., style and material). Furthermore, their complex training pipelines and closed-source code pose significant challenges for reproducibility. Disco-LoRA pioneers the exploration of this limitation, successfully achieving customized video generation across diverse concepts—a task that even current commercial models cannot fully accomplish. (3) Moreover, our method has strong generalizability. Although our experiments are exclusively conducted on WAN2.1-1.3B, analyzing the distribution trends of LoRA weights across DiT layers and applying Z-Score Statistical Regularization allows our approach to be effectively adapted to different base models.

\noindent\textbf{Future Work.} To address these limitations and advance the field, our future research will focus on two key directions. First, we aim to push the boundaries of compositional generation by scaling Disco-LoRA to handle a larger number of concurrent concepts, rigorously testing its stability and disentanglement capabilities in high-complexity scenarios. Second, we plan to broaden the scope of disentanglement beyond object-centric attributes. We intend to incorporate additional dimensions such as background synthesis and lighting control into our framework, thereby achieving a more comprehensive and fine-grained control over the entire video generation process.

\begin{figure*}[t]
    \centering
    \includegraphics[width=0.9\textwidth]{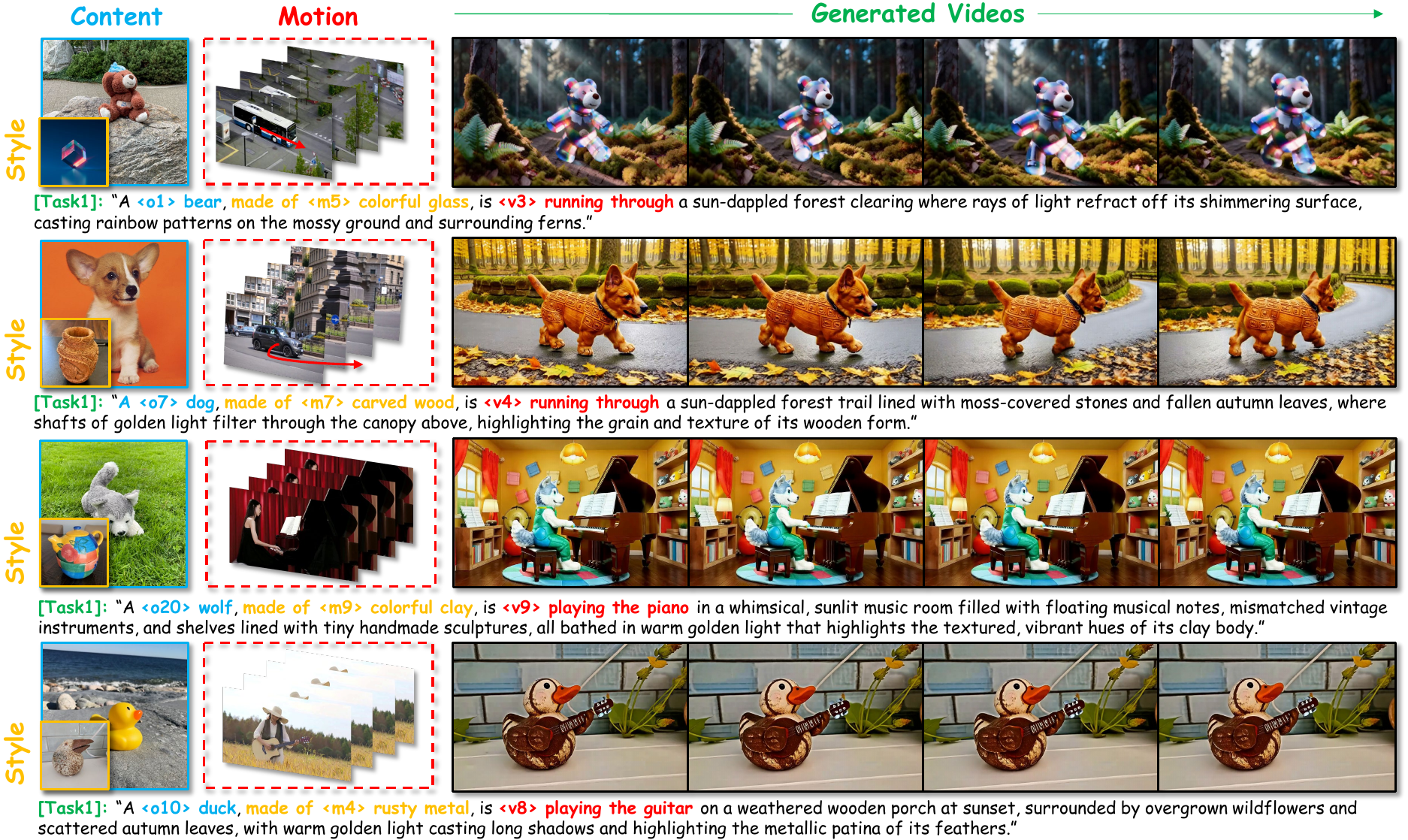}
    \caption{Additional qualitative results for Task 1. We generated a diverse set of customized videos to validate our approach. These samples highlight the accurate extraction of target concepts and the robust generalization of our model.}
    \label{fig_moretasks1}
\end{figure*}
\begin{figure*}[t]
    \centering
    \includegraphics[width=0.9\textwidth]{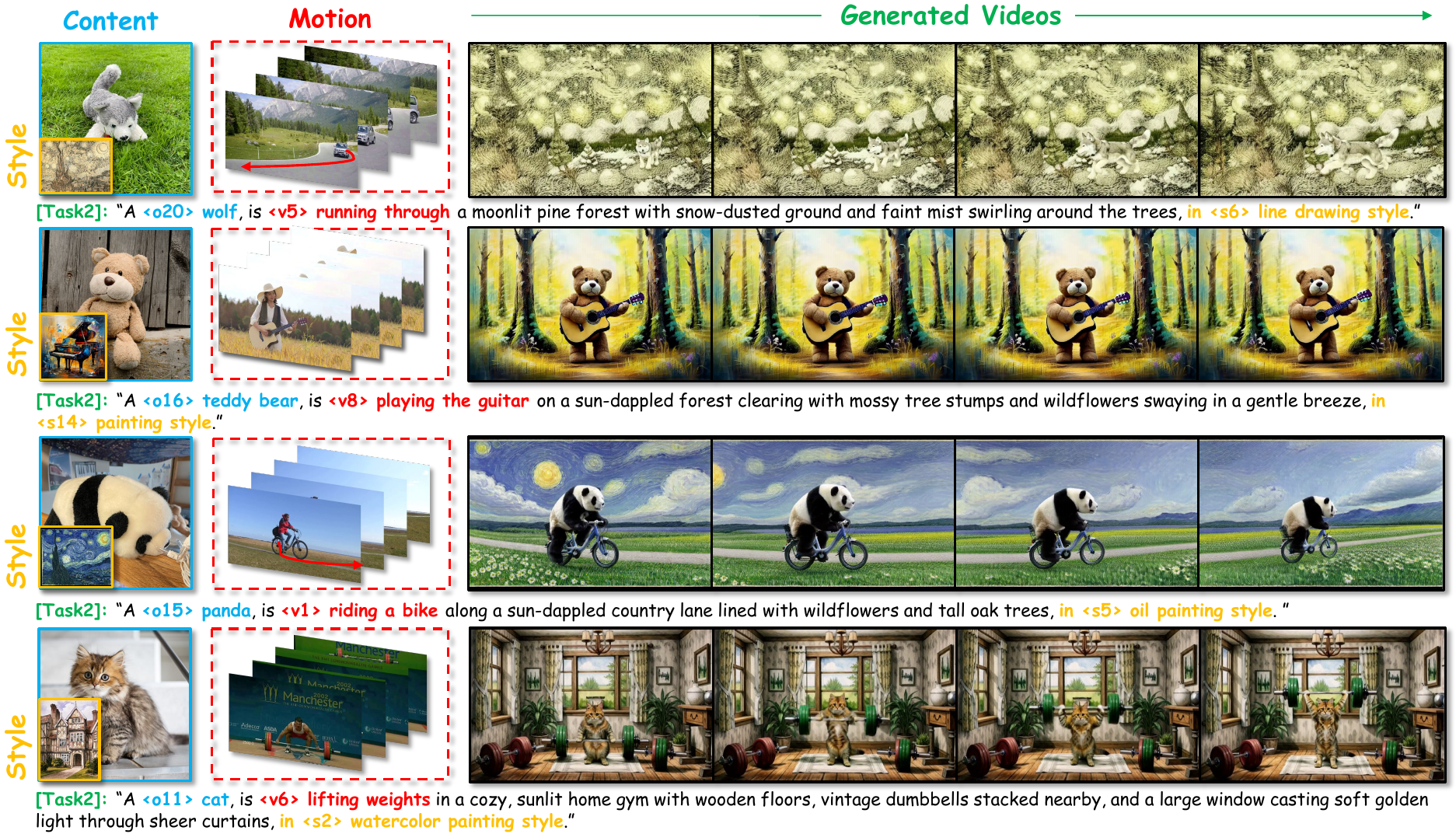}
     \caption{Additional qualitative results for Task 2. We generated a diverse set of customized videos to validate our approach. These samples highlight the accurate extraction of target concepts and the robust generalization of our model.}
    \label{fig_moretasks2}
\end{figure*}
\begin{figure*}[t]
    \centering
    \includegraphics[width=0.9\textwidth]{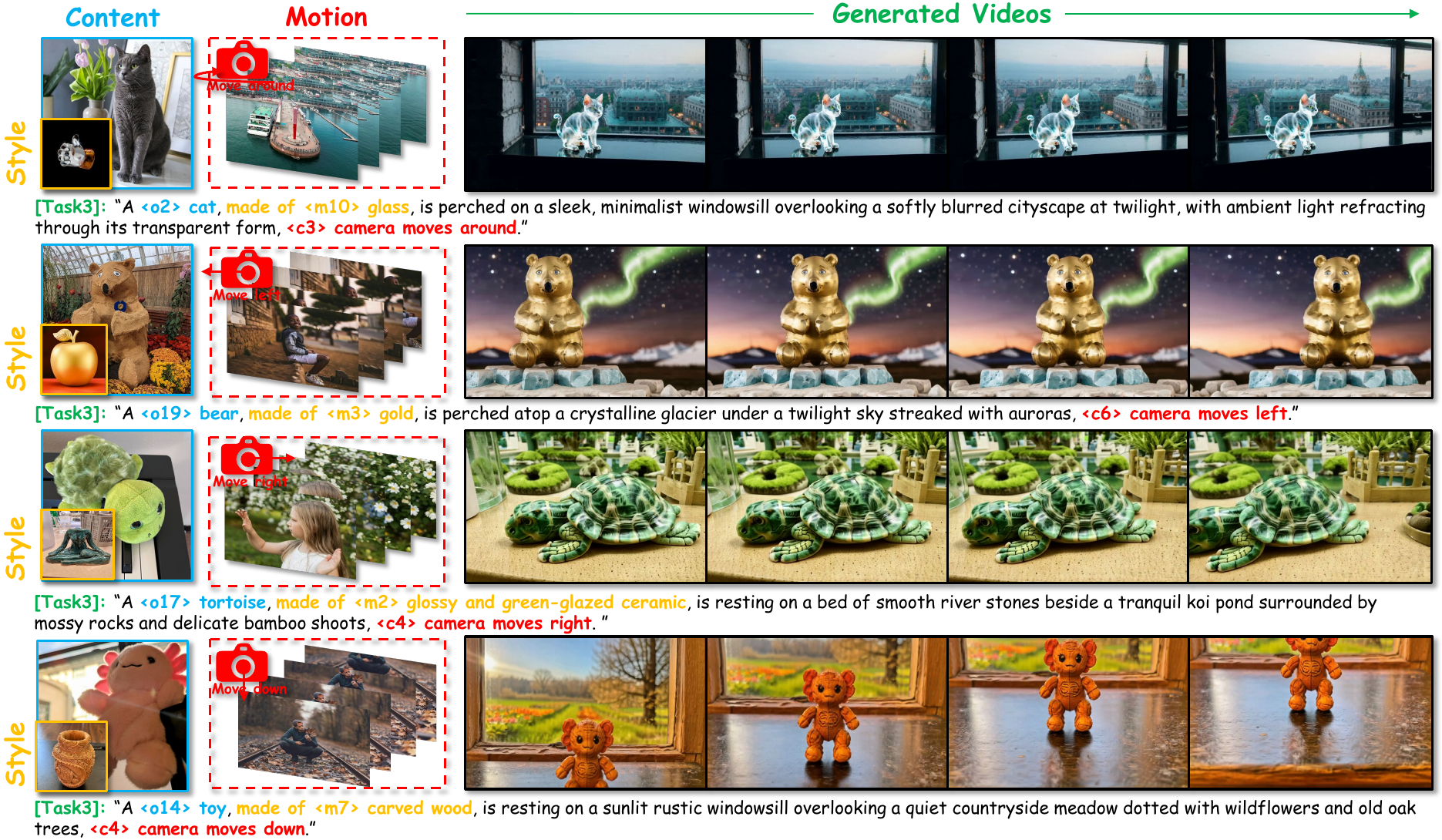}
     \caption{Additional qualitative results for Task 3. We generated a diverse set of customized videos to validate our approach. These samples highlight the accurate extraction of target concepts and the robust generalization of our model.}
    \label{fig_moretasks3}
\end{figure*}
\begin{figure*}[t]
    \centering
    \includegraphics[width=0.9\textwidth]{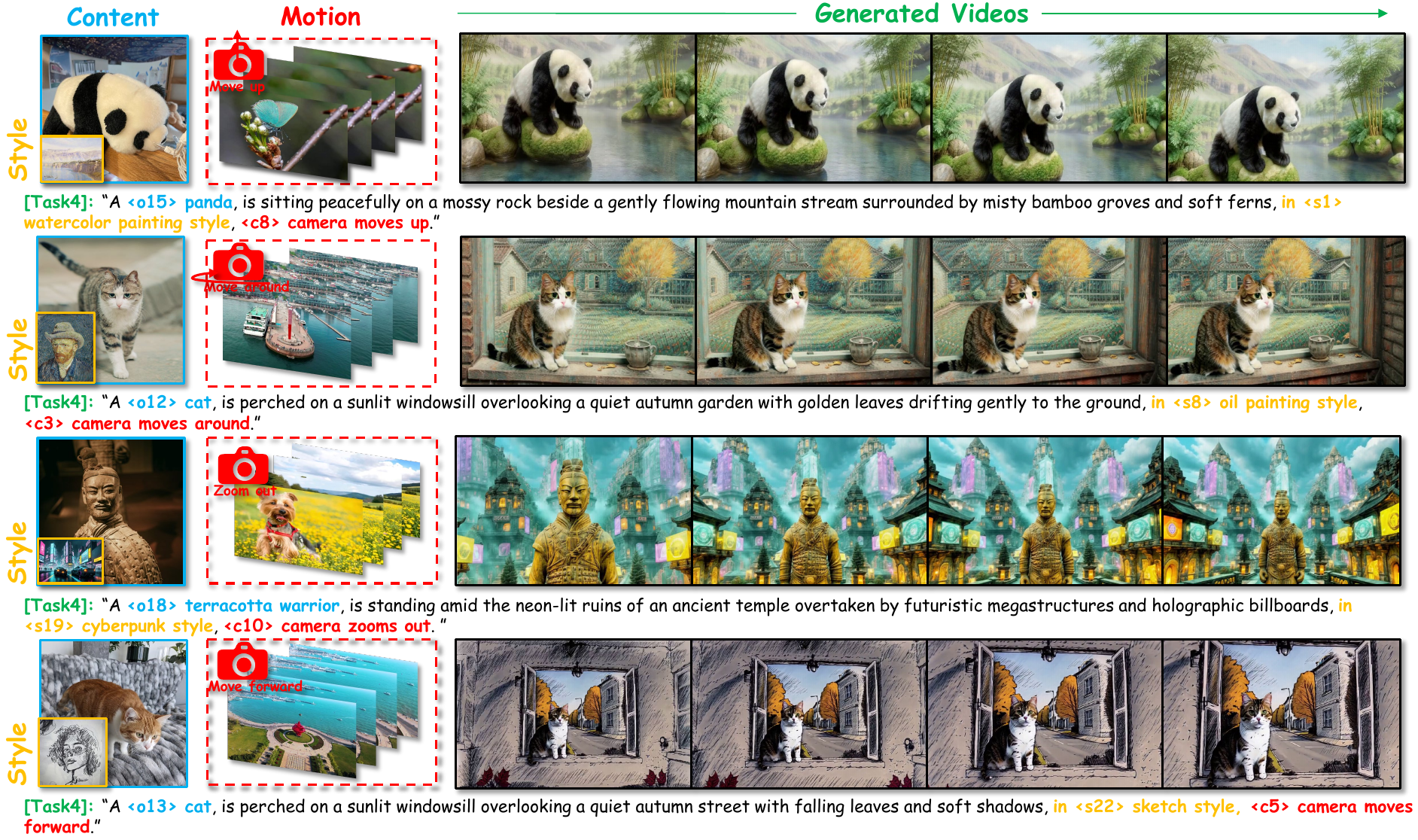}
    \caption{Additional qualitative results for Task 4. We generated a diverse set of customized videos to validate our approach. These samples highlight the accurate extraction of target concepts and the robust generalization of our model.}
    \label{fig_moretasks4}
\end{figure*}

\end{document}